\documentclass[acmsmall]{acmart}

\usepackage{lineno}   
\usepackage{booktabs} 
\usepackage{url}  

\usepackage{multirow}  
\usepackage{color}  
\newtheorem{upper bound}{Upper bound}

\usepackage{diagbox}
\usepackage{amsmath}
\usepackage{arydshln} 
\usepackage{color} 
\usepackage{booktabs} 
\usepackage{multirow} 
\usepackage{makecell} 

\usepackage{graphicx}
\usepackage{algorithmic}
\usepackage{xcolor}
\usepackage{xspace}
\usepackage{flushend}
\usepackage{enumitem}
\usepackage{pifont}
\usepackage{subfloat}
\usepackage{subcaption}

\usepackage[linesnumbered,ruled]{algorithm2e} 

\SetAlFnt{\small}
\SetAlCapFnt{\small}
\SetAlCapNameFnt{\small}
\SetAlCapHSkip{0pt}
\IncMargin{-\parindent}

\usepackage{amsthm}

\newtheorem*{theorem*}{Theorem}

\usepackage{xspace}
\usepackage{multirow}

\setcopyright{rightsretained}
\acmJournal{JACM}

\begin{document}
	
\title{Enhanced Pre-training of Graph Neural Networks for Million-Scale Heterogeneous Graphs}

\author{Shengyin~Sun}
\affiliation{
	\institution{City University of Hong Kong}
	\city{Hong Kong SAR}
	\country{China}
}
\email{shengysun4-c@my.cityu.edu.hk}

\author{Chen~Ma}
\authornote{This is the corresponding author.}
\affiliation{
	\institution{City University of Hong Kong}
	\city{Hong Kong SAR}
	\country{China}
}
\email{chenma@cityu.edu.hk}

\author{Jiehao~Chen}
\affiliation{%
	\institution{China Academy of Industrial Internet}
	\city{Beijing}
	\country{China}
}
\email{chenjiehao@china-aii.com}


\begin{abstract}

In recent years, graph neural networks (GNNs) have facilitated the development of graph data mining. However, training GNNs requires sufficient labeled task-specific data, which is expensive and sometimes unavailable. To be less dependent on labeled data, recent studies propose to pre-train GNNs in a self-supervised manner and then apply the pre-trained GNNs to downstream tasks with limited labeled data. However, most existing methods are designed solely for homogeneous graphs (real-world graphs are mostly heterogeneous) and do not consider semantic mismatch (the semantic difference between the original data and the ideal data containing more transferable semantic information). In this paper, we propose an effective framework to pre-train GNNs on the large-scale heterogeneous graph. We first design a structure-aware pre-training task, which aims to capture structural properties in heterogeneous graphs. Then,  we design a semantic-aware pre-training task to tackle the mismatch. Specifically, we construct a perturbation subspace composed of semantic neighbors to help deal with the semantic mismatch. Semantic neighbors make the model focus more on the general knowledge in the semantic space, which in turn assists the model in learning knowledge with better transferability. Finally, extensive experiments are conducted on real-world large-scale heterogeneous graphs to demonstrate the superiority of the proposed method over state-of-the-art baselines. Code available at \url{https://github.com/sunshy-1/PHE}.

\end{abstract}

\keywords{Graph neural networks, Large-scale heterogeneous graphs, Pre-training techniques, Contrastive learning.}

\authorsaddresses{\textbf{Authors' addresses}: 
Shengyin Sun, City University of Hong Kong, Hong Kong SAR, China, shengysun4-c@my.cityu.edu.hk; Chen Ma, City University of Hong Kong, Hong Kong SAR, China, chenma@cityu.edu.hk; Jiehao Chen,  China Academy of Industrial Internet, Beijing, China, chenjiehao@china-aii.com.
}

\ccsdesc[500]{Information systems~Data mining}

\renewcommand\shortauthors{S. Sun et al.}

\maketitle

\section{Introduction}
\label{ssy1210:introduction}
{A}{s} powerful tools on learning graph-structured data~\cite{DBLP:conf/icml/TangLi22, arxiv:SunRen23},
Graph Neural Networks (GNNs) have been widely used and demonstrated effective in real-world scenarios, including but not limited to recommendation systems~\cite{DBLP:conf/kdd/YingHe18,DBLP:conf/ecml/SunMa24}, traffic-flow forecasting~\cite{DBLP:conf/aaai/ChenLi19,arxiv:XuLiu25} and drug discovery \cite{arxiv:SunYu25}. The capacities of GNNs have shifted the focus of graph data mining from structural feature engineering to representation learning. The core idea of GNNs is to learn representations by recursively aggregating messages from neighbor nodes, which enables both structure and content information to be preserved. It is worth noting that GNNs are usually trained in a supervised manner, which requires a considerable amount of labeled data for different tasks. However, the acquisition of large amounts of labeled data is expensive and sometimes impossible in practical applications. The effective utilization of information within unlabeled data has thus become a highly significant issue.

Inspired by the development of pre-training techniques in computer vision and natural language processing, recent studies propose to apply pre-training strategies on graphs to alleviate GNNs' reliance on task-specific labeled data~\cite{arxiv:HuLiu19,DBLP:conf/aaai/LuJiang21,DBLP:conf/prcv/WangWang24,arxiv:WangNie25,DBLP:journals/JEAI/LiuZhang25,DBLP:journals/pr/SunLiu23}. The goal of pre-training strategies on graphs is to capture transferable knowledge (i.e., intrinsic properties of graphs) from unlabeled data (e.g., unlabeled graph structures), which can be flexibly applied to various downstream tasks. With the help of obtained transferable knowledge, GNNs in downstream tasks can achieve competitive performance by using only a small amount of labeled data. Most of the existing methods~\cite{arxiv:ThomasMax16,arxiv:HuLiu19} adopt the learning framework of \emph{pre-training first, then fine-tuning}. Specifically, they first pre-train a GNN by completing pretext tasks, and then fine-tune the pre-trained GNN on downstream tasks. It is worth noting that the difference between the pre-training strategies mainly lies in the pretext tasks. For example, the pretext task in the Variational Graph Auto-Encoder (GAE)~\cite{arxiv:ThomasMax16} aims at graph reconstruction while the pretext task in the Deep Graph Infomax (DGI)~\cite{DBLP:conf/iclr/VelickovicFedus19} aims at maximizing the mutual information between embeddings at different levels. 

Existing pre-training strategies have achieved satisfactory results in many graph data mining tasks, however, most of them are designed for homogeneous graphs (graphs with only one type of nodes/edges). It should be noted that graphs in real-world applications are mostly heterogeneous graphs, which contain different types of nodes/edges. For example, academic graphs contain different types of nodes such as ``author'' and ``paper'', and also contain different types of edges such as ``write'' and ``cite''. Considering the uniqueness of heterogeneous graphs, some recent studies attempt to design pre-training strategies by using semantic patterns~\cite{DBLP:conf/VLDB/SunHan11} such as the meta-path and the network-schema\footnote{The instance of network-schema contains all types of nodes and edges in the graph~\cite{DBLP:conf/VLDB/SunHan11}, which can be regarded as a template for the heterogeneous graph.}, which hopefully capture the rich structural and semantic information in heterogeneous graphs~\cite{DBLP:conf/kdd/JiangJia21}. 
\begin{figure}[h]
    \centering
    \includegraphics[scale=0.5]{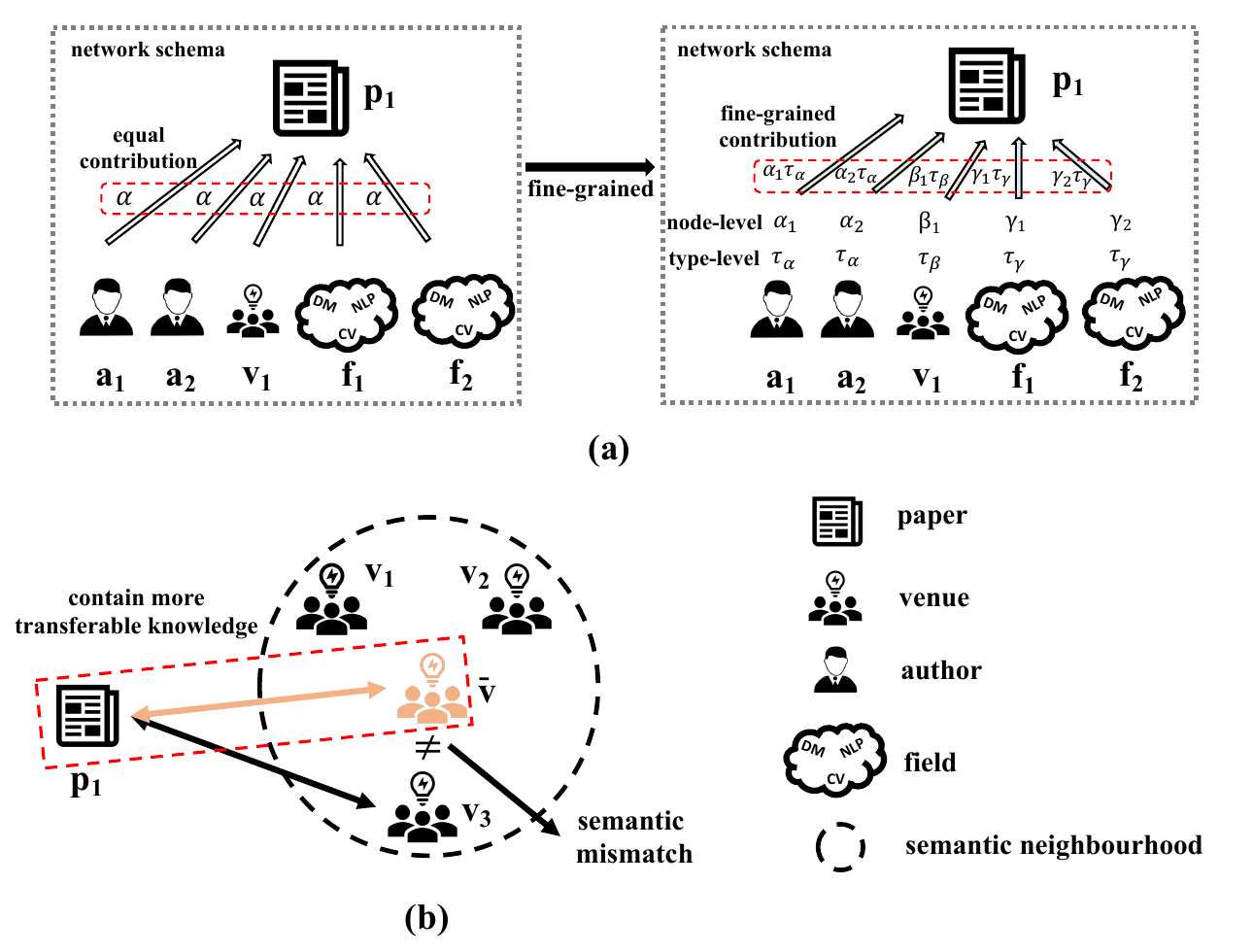}
    \caption{Shortcomings of existing methods. (a) The fine-grained contributions are not considered. (b) The semantic mismatch is not considered.}
    \label{ssy1210:two-problem}
\end{figure}

Although existing methods have achieved satisfactory results to some extent, we argue that there are still avenues to improve. First, methods based on semantic patterns~\cite{DBLP:conf/kdd/JiangJia21,DBLP:conf/cikm/JiangLu21} do not consider fine-grained heterogeneous information, that is, they assume that each node in the semantic pattern contributes equally to capturing heterogeneous information. \textcolor{black}{However, we argue that a more reasonable assumption would be that different types of nodes contribute in varying degrees. 
More specifically, in the process of constructing contrastive views for pre-training, we should model the fine-grained heterogeneity within heterogeneous graphs from both type-specific and node-specific perspectives. In other words, the weights of different types of nodes should be distinguished, and individual nodes of the same type should also be differentiated. A motivating example under the network schema is shown in Fig. \ref{ssy1210:two-problem}(a).} Second, existing studies do not consider semantic mismatch~\cite{DBLP:conf/iclr/VelickovicFedus19,DBLP:conf/www/XiaWu22}. The so-called semantic mismatch refers to the semantic-level difference between the original data (e.g., $v_3$ in Fig. \ref{ssy1210:two-problem}(b)) and the ideal data (data containing more transferable semantic information, e.g., $\bar{v}$ in Fig. \ref{ssy1210:two-problem}(b)). An illustration of the semantic mismatch is shown in Fig. \ref{ssy1210:two-problem}(b). Assume that the original data is ``$p_1$ is published at $v_3$''. Compared with the knowledge of ``$p_1$ is published at $v_3$'', a more transferable knowledge is  ``$p_1$ can be published at venues similar to $v_3$'' (for example, it is reasonable for a paper in the field of computer vision to be published at CVPR, and it is also reasonable to publish this paper at ICCV). As shown in Fig. \ref{ssy1210:two-problem}(b), an average embedding $\bar{{{v}}}$ that considers semantic neighbors seems to be more helpful for the model to learn transferable knowledge. In other words, paying attention to the semantic neighbors of nodes is conducive to the model learning knowledge with better transferability. 
\textcolor{black}{Moreover, some existing methods face challenges when extending to large-scale heterogeneous graphs (large-scale heterogeneous graphs are more closely aligned with real-world applications). For example, the SHGP proposed in \cite{DBLP:conf/nips/YangGuan22} requires label propagation over the entire heterogeneous graph to obtain high-quality pseudo-labels for guiding the pre-training process. However, the scale of heterogeneous graphs in the real world includes millions of nodes, making label propagation on such large-scale graphs impossible.}

To address the above shortcomings, we propose the \emph{\underline{P}re-training Graph Neural Networks on Large-Scale \underline{H}eterogeneous Graphs with \underline{E}nhancement (PHE)}. Inspired by contrastive learning, contrastive pre-training strategies are designed to capture heterogeneous information and cope with mismatched situations. Specifically, we design two pre-training strategies according to the typical triplet-like structure (\textless query, positive, negative\textgreater) in contrastive learning, namely structure-aware pre-training with enhanced query samples and semantic-aware pre-training with enhanced positive/negative samples, respectively. In the structure-aware pre-training task, a network-schema subspace is constructed to help capture the heterogeneous structure, where the columns of the network-schema subspace are the embeddings of nodes in the network schema. Moreover, the attention mechanism is applied to help the model capture fine-grained heterogeneous information. In particular, we exploit attention scores to measure the contribution of different nodes to model the heterogeneous information. In the semantic-aware pre-training task, a perturbation subspace is designed to help handle mismatched situations, where the columns of the perturbation subspace are the embeddings of semantic neighbors. \textcolor{black}{We further demonstrate how to adapt the proposed pre-training strategy to operate in a mini-batch manner, making it fit into the large-scale heterogeneous graphs encountered in practical applications.} Experimental results show that the proposed method outperforms its counterparts and achieves performance improvements on large-scale heterogeneous graphs. The contributions of this paper are summarized below: 
\begin{itemize}[leftmargin=*]
\item{{We investigate pre-training strategies for heterogeneous graphs at scale, focusing on the design of efficient and effective frameworks applicable to graphs with up to millions of nodes and tens of millions of edges. Our approach addresses the challenges of scalability and resource efficiency, enabling practical deployment in large-scale settings.}}
\item{{We design two pre-training tasks for large-scale heterogeneous graphs. To capture rich structural information in heterogeneous graphs, we design a contrastive structure-aware pre-training task. Specifically, we construct a schema-network subspace and employ an attention mechanism to enable the model to capture fine-grained heterogeneous patterns. To deal with the problem of semantic mismatch, we design the contrastive semantic-aware pre-training task. In this task, we construct a perturbation subspace to help the model pay attention to semantic neighbors, which is beneficial for the model to learn knowledge with better transferability.}}
\item{{Extensive experiments are conducted on large-scale heterogeneous graphs to evaluate the performance of the proposed method PHE. The results demonstrate that PHE significantly outperforms state-of-the-art approaches, validating the effectiveness of the proposed pre-training strategies.}}
\end{itemize}

\section{Related Work}
\label{sec:relatedwork}

In this section, we summarize and discuss work related to the proposed method.

\textbf{Graph Neural Networks}. Neural networks have revolutionized many fields~\cite{arxiv:ZengChang25,arxiv:ZengChang24,DBLP:journals/IJSTAR/GuoChen25,DBLP:journals/RS/GuoChen22,DBLP:journals/IJSTAR/TianWang24}. However, they cannot be directly applied to non-Euclidean data (e.g., graph). To solve this problem, graph neural networks came into being. One of the most famous GNN architectures is the graph convolutional network (GCN) proposed by Kipf \emph{et al.}~\cite{DBLP:conf/iclr/Thomas16}, which aggregates the feature information of each node's one-hop neighbors. Many other famous architectures can be seen in ~\cite{DBLP:conf/nips/HamiltonYing17,DBLP:conf/iclr/VelickovicCucurull18}. It should be pointed out that the GNNs mentioned above are all designed for homogeneous graphs. Considering the widespread existence of heterogeneous graphs, many studies have been carried out recently to extend GNN to heterogeneous graphs. In \cite{arxiv:SchlichtkrullKipf17}, the relational graph convolutional network (R-GCN) is developed to handle graphs with multiple types of edges. In~\cite{DBLP:conf/www/WangJi21}, the heterogeneous graph attention network (HAN) is proposed by Wang~\emph{et al.}, which aims to capture semantic information in heterogeneous graphs by using the hierarchical attention. In ~\cite{DBLP:conf/www/HuDong20}, Hu~\emph{et al.} design the heterogeneous graph transformer (HGT) architecture, which aims to model heterogeneity by applying the transformer model. The GNN is the cornerstone of our framework.

\textbf{Contrastive Learning}. In recent years, contrastive learning has become a research hotspot in the field of unsupervised learning~\cite{DBLP:conf/cvpr/YuanLin21,DBLP:journals/TGRS/TaoLi23}. Various frameworks for contrastive learning have been proposed to solve problems in different fields, such as the SimCLR in the field of computer vision~\cite{DBLP:conf/icml/ChenHinton20} and the DeCLUTR in the field of natural language processing~\cite{DBLP:conf/ACL/GiorgiNitski21}. The common goal of these frameworks is to learn such an embedding space where similar pairs of samples are close to each other and dissimilar pairs are far apart. Inspired by the research on contrastive learning in the fields of computer vision and natural language processing, a lot of work has been carried out in recent years to study the application of contrastive learning in the field of graph learning. In \cite{DBLP:conf/aaai/zengxie21}, Zeng \emph{et al.} introduce contrastive learning into the graph classification task to address the overfitting problem. In \cite{DBLP:journals/tkde/ZhengJin21}, Zheng \emph{et al.} propose the SL-GAD, which applies the contrastive learning paradigm to the graph anomaly detection task. More specifically, the SL-GAD first constructs different contextual subgraphs (views) and then performs multi-view contrastive learning. Multiple subgraphs allow the model to explore richer structure information, thereby enabling more effective capture of anomalies in the structure space. In this paper, contrastive learning is regarded as a powerful tool to help the proposed model capture high-order information in heterogeneous graphs.

\textbf{Heterogeneous Graphs. } Heterogeneous graphs, which incorporate different types of entities (i.e., nodes) and relations into a network~\cite{DBLP:journals/tbd/WangBo22,DBLP:journals/CAGD/WangZhang24}, have become pervasive across a multitude of real-world scenarios, including bibliographic networks~\cite{DBLP:journals/kbs/XuZhong22}, social networks~\cite{DBLP:journals/tii/WangSun20}, and recommendation systems~\cite{DBLP:journals/tkde/ShiHu18}. The diverse relationships within heterogeneous graphs enable them to embrace the rich semantic and structural information present in real-world data. In order to extract rich information from heterogeneous graphs, learning embeddings for heterogeneous graphs has become a significant research topic. Some early work use matrix factorization (MF) methods~\cite{DBLP:conf/PANS/Newman06,DBLP:journals/NTI/ShervashidzeSchweitzer11} to generate latent features for heterogeneous graphs. However, these MF-based methods are difficult to scale to large-scale heterogeneous graphs due to the high time complexity of decomposing large-scale matrices. To address this challenge, many studies have attempted to combine graph representation learning methods with heterogeneous graphs. For example, Perozzi \emph{et al.} introduce DeepWalk in~\cite{DBLP:conf/kdd/PerozziAl-Rfou14}, which feeds a set of short random walks over the graph into the SkipGram~\cite{DBLP:conf/nips/MikolovSutskever13} to approximate the node co-occurrence probability within those walks and obtain node embeddings. However, ``shallow models'' like Deepwalk are unable to effectively capture complex patterns in heterogeneous graphs. Inspired by the powerful capability of GNNs in modeling complex relations in graphs, designing specialized GNN architectures for heterogeneous graphs has emerged as a mainstream~\cite{DBLP:conf/www/WangJi21,DBLP:conf/www/HuDong20}. 

\textbf{Graph Pre-training}. Inspired by the development of pre-training techniques in computer vision \cite{DBLP:conf/icml/ChenHinton20, DBLP:conf/cvpr/YuanLin21,DBLP:journals/TCSVT/ZhangChen23} and natural language processing \cite{DBLP:conf/NAACL/DevlinChang19}, many studies have been conducted to design pre-training strategies on unlabeled graphs~\cite{DBLP:conf/nips/YouChen20,DBLP:conf/www/XiaWu22,DBLP:conf/kdd/WangLiu18}. The goal of graph pre-training is to learn generic initialization parameters for GNNs in a self-supervised manner. With more and more attention paid to GNNs, exploring how to use unannotated data to pre-train GNNs has gradually become a hot topic in recent years. Kipf \emph{et al.} propose the GAE in \cite{arxiv:ThomasMax16}. The GAE aims to pre-train the GNN by completing the task of graph reconstruction. In \cite{DBLP:conf/nips/HamiltonYing17}, Hamilton \emph{et al.} introduce the unsupervised GraphSAGE, which pre-trains GNNs by optimizing a loss that encourages structurally adjacent nodes to have similar embeddings. Velickovic \emph{et al.} propose the DGI in \cite{DBLP:conf/iclr/VelickovicFedus19}, which maximizes the mutual information between graph summary embeddings and local node embeddings. The pre-training strategies mentioned above are all designed for homogeneous graphs. However, heterogeneous graphs are more common in practice. In \cite{DBLP:conf/kdd/JiangJia21}, a framework of pre-training GNNs on the heterogeneous graph (PTHGNN) is proposed, which aims to contrastively preserve properties in heterogeneous graphs by using contrastive learning.
Motivated by these existing works, we design a novel contrastive framework in this paper, which is expected to simultaneously capture the heterogeneous information and deal with the problem of semantic mismatch.

\section{Preliminary}
\subsection{Notations}
We follow the common manner that vectors and matrices are denoted by bold lower case letters (e.g., $\mathbf{{v}}$) and bold upper case letters (e.g., $\mathbf{{R}}$), respectively. Sets are denoted by calligraphic letters (e.g., $\mathcal{A}$), $\vert \cdot \vert$ represents the number of elements in the set (e.g., $\vert\mathcal{A}\vert$). Superscript $(\cdot)^{\top}$ denotes transpose. $\mathbb{R}^{m\times n}$ is real matrix space of dimension $m\times n$. $\|$ is the concatenation operation. $\odot$ denotes the element-wise product of vectors.

\subsection{Heterogeneous Graph}
A heterogeneous graph is a graph with multiple types of nodes and edges. Define a heterogeneous graph as $\mathcal{G}=\{\mathcal{V}, \mathcal{E}, \mathbf{{A}}, \mathbf{{X}}\}$, where $\mathcal{V}$ is the set of nodes, $\mathcal{E}$ is the set of edges, $\mathbf{{A}}\in \mathbb{R}^{\vert\mathcal{V}\vert\times \vert\mathcal{V}\vert}$ is an adjacency matrix, and $\mathbf{{X}} \in \mathbb{R}^{\vert\mathcal{V}\vert\times d}$ is a node feature matrix (the dimension of the feature is $d$). Considering multiple types of nodes and edges in a heterogeneous graph $\mathcal{G}$, two mapping functions are used to map each node/edge to its corresponding type, i.e., a node-type mapping function $f_{\mathcal{V}}:\mathcal{V}\rightarrow\mathcal{P}$ and an edge-type mapping function 
$f_{\mathcal{E}}:\mathcal{E}\rightarrow\mathcal{Q}$, where $\mathcal{P}$ and $\mathcal{Q}$ are labels of edges and nodes, respectively. Note that $|\mathcal{P}|+|\mathcal{Q}|>2$ is satisfied in the heterogeneous graph $\mathcal{G}$. {The network schema is a widely used concept in heterogeneous graphs, which can be viewed as a meta-template of $\mathcal{G}$. It is a directed graph defined over node types $\mathcal{P}$, with edges representing relations from $\mathcal{Q}$. An intuitive example of the network schema is given in Fig. \ref{ssy1210:framework} \#2), which includes all the node types and relation types. The network schema describes the direct connections between different nodes, which naturally represents the local structure.}

\subsection{Graph Neural Network}
In recent years, GNNs have achieved remarkable success in the field of graph data mining. GNNs can obtain high-quality node representations by using the message-passing strategy. Formally, the representation of node $v$ at the $k$-th layer ${\mathbf{{h}}}_{v}^{k}$ can be written as:
\begin{equation}
    \label{ssy1210:gnneq}
    {\mathbf{{h}}}_{v}^{k}={\textsc{AGGREGATE}}_{\boldsymbol{\Theta}_{k}}\left(\left\{{\mathbf{{h}}}_{u}^{k-1}, u\in\mathcal{N}_{v}\cup\left\{v\right\}\right\}\right),
\end{equation}
where $\mathcal{N}_{v}$ is the set of neighbors of node $v$, ${\textsc{AGGREGATE}}$ is the operation of aggregating messages, and ${\boldsymbol{\Theta}_{k}}$ is a matrix of learnable parameters in the $k$-th layer. For the sake of brevity, we define the final node representation of node $v$ obtained from a k--layer GNN as ${\mathbf{{h}}}_{v}=f_{\boldsymbol\Theta}\left(v,\mathcal{G}\right)$, where $\boldsymbol\Theta=\left[\boldsymbol{\Theta}_{1},\cdots,\boldsymbol{\Theta}_{k}\right]$ is the concatenated learnable parameters and $f$ is an abstract parameterized function containing all operations (i.e., $\textsc{AGGREGATE}$) in the GNN.

\begin{table}
\renewcommand\arraystretch{1.0}
  \centering
  \caption{The Important and Frequently Used Notations.}
  \scalebox{1.0}{
    \begin{tabular}{ll}
    \toprule[1.5pt]
    \textbf{Notation}     & \textbf{Explanation} \\
    \midrule
    $\mathcal{G}$     & A heterogeneous graph.\\\addlinespace[0.25em]
    
    $\boldsymbol{\Theta}$     & A learnable parameter tensor in a GNN.\\\addlinespace[0.25em]
    
    \makecell[l]{$f_{\boldsymbol{\Theta}}$}     & \makecell[l]{The abstraction function of a GNN with the \\ parameter tensor $\boldsymbol{\Theta}$.}\\\addlinespace[0.25em]
    
    \makecell[l]{$f_{\mathcal{V}}$}     & \makecell[l]{A node-type mapping function (input a node, \\output its type).}\\\addlinespace[0.25em]
    
    \makecell[l]{$f_{\mathcal{E}}$}     & \makecell[l]{An edge-type mapping function (input an edge, \\output its type).}\\\addlinespace[0.25em]
    
    $\mathbf{h}_{v}$     & The representation of node $v$.\\\addlinespace[0.25em]
    
    $\mathbf{S}_{1}$     & The network-schema subspace.\\\addlinespace[0.25em]
    
    $\mathbf{S}_{2}$     & The perturbation subspace. \\\addlinespace[0.25em]
    
    $\mathbf{p}_{a}^{v}$     & The node-level weight vector for node $v$. \\\addlinespace[0.25em]
    
    $\mathbf{p}_{b}^{v}$     & The type-level weight vector for node $v$.\\\addlinespace[0.25em]
    
    $\mathbb{R}^{d}$     & A $d$-dimensional real space\\\addlinespace[0.25em]
    
    \makecell[l]{$\alpha_{v,u_{i}}$}     & \makecell[l]{The contribution of the node $u_{i}$ to node $v$ within\\ node type $f_{\mathcal{V}}(u_{i})$.} \\\addlinespace[0.25em]
    
    \makecell[l]{$\beta_{v,\varpi_{k}}$} & \makecell[l]{The contribution of the node type $\varpi_{k}$ to node $v$.} \\\addlinespace[0.25em]
    
    $\mathcal{S}_{\left( v, r, u\right)_{t}}^{1}$     & \makecell[l]{A set of negative sampling nodes for the structure-\\aware pre-training task (current bacth $t$) based on \\the relationship $r$ between node $v$ and node $u$.} \\\addlinespace[0.25em]
    $\mathcal{S}^{1}_{Queue}$     & \makecell[l]{The negative sample sets from a dynamic queue in \\the structure-aware pre-training task.}\\\addlinespace[0.25em]
    $\mathcal{S}^{(1)}$     & \makecell[l]{The final set of negative samples in the structure-\\aware pre-training task.}\\\addlinespace[0.25em]
    $\mathcal{S}_{\left( v, r, u\right)_{t}}^{2}$     & \makecell[l]{A set of negative sampling nodes for the semantic-\\aware pre-training task (current bacth $t$) based on \\the relationship $r$ between node $v$ and node $u$.} \\\addlinespace[0.25em]
    $\mathcal{S}^{2}_{Queue}$     & \makecell[l]{The negative sample sets from a dynamic queue in \\the semantic-aware pre-training task.}\\\addlinespace[0.25em]
    $\mathcal{S}^{(2)}$     & \makecell[l]{The final set of negative samples in the semantic-\\aware pre-training task.}\\\addlinespace[0.25em]
    \bottomrule[1.5pt]
    \end{tabular}}
  \label{ssy0507:notations}
\end{table}%
\section{Methodology}
\label{ssy1210:method}
\begin{figure*}
    \centering
    \includegraphics[scale=0.41]{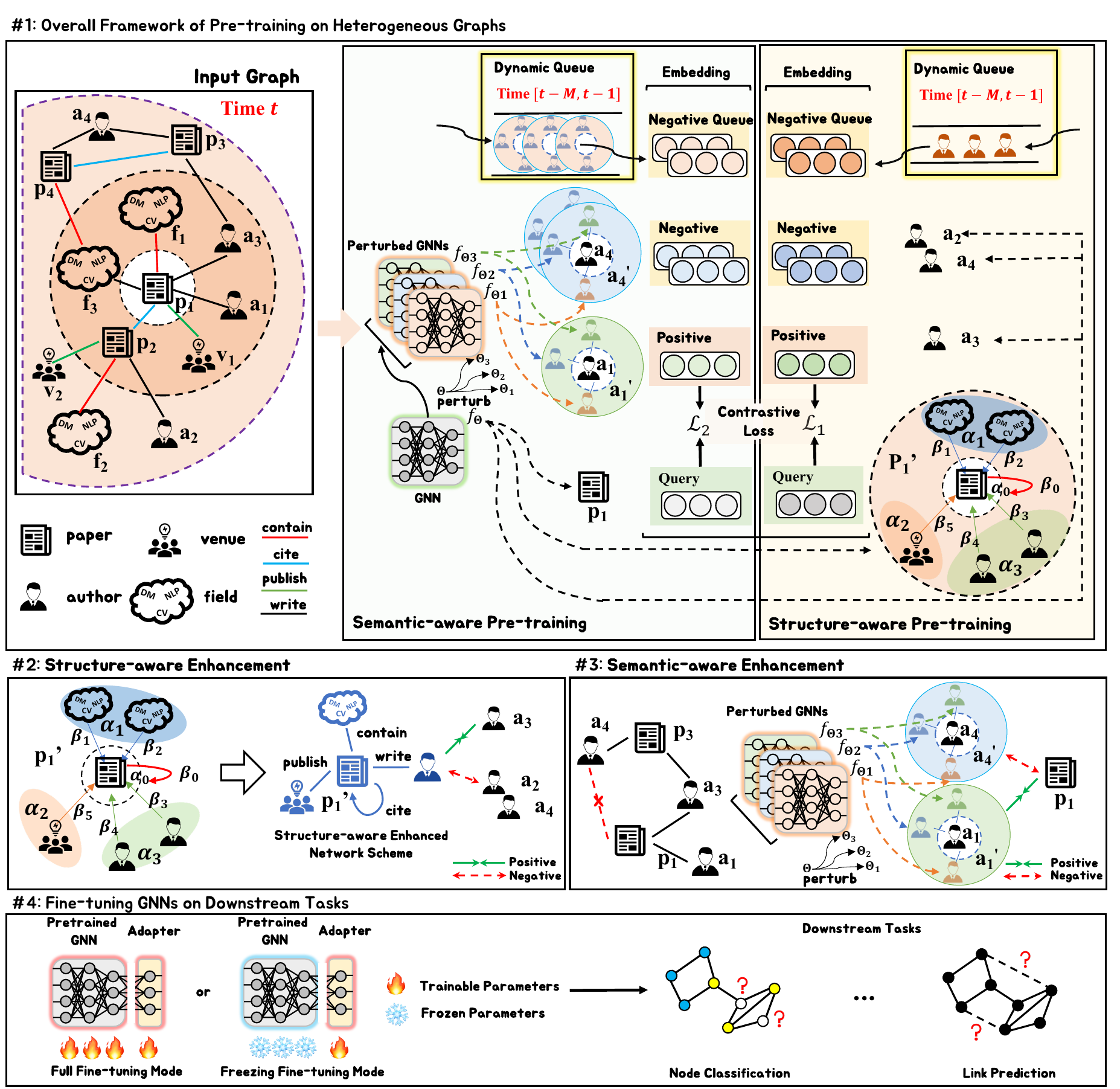}
    \caption{The demonstration of the proposed model. \#1) The overall framework includes two components: the structure-aware pre-training task and the semantic-aware pre-training task. In the structure-aware pre-training task, enhanced query samples are used to capture high-order properties of the heterogeneous
graph. In semantic-aware pre-training task, enhanced positive/negative samples are used to guide the GNN to learn more transferable knowledge. {\#2) and \#3) describe the proposed modules. The module in \#2) utilizes a structure-enhanced network schema for contrastive learning to capture fine-grained heterogeneous information. The module in \#3) constructs positive/negative pairs through parameter perturbation to address the issue of semantic mismatch.} \#4) After obtaining the pre-trained GNN, we fine-tune it on the downstream tasks. There are two modes for fine-tuning, namely full fine-tuning mode and frozen fine-tuning mode. The full fine-tuning mode is to fine-tune both the pre-trained GNN and the downstream task adapter (e.g., the classification head tailored for the downstream node classification task) simultaneously, while the frozen fine-tuning mode is to freeze the parameters of pre-trained GNN and only fine-tune the parameters of the adapter.}
    \label{ssy1210:framework}
\end{figure*}

In this section, a pre-training framework is designed for GNNs on the heterogeneous graph. First, we introduce the general pipeline for pre-training on heterogeneous graphs. Second, we design pre-training strategies for large-scale heterogeneous graphs, including the Transformer-based heterogeneous encoding, the design of structure-aware pre-training task, the design of semantic-aware pre-training task, and the techniques for applying the proposed pre-training strategies to large-scale heterogeneous graphs. Finally, loss functions in the propsoed two tasks are jointly optimized. The overview of the proposed method is shown in Fig. \ref{ssy1210:framework}.

\subsection{Pre-Training for Heterogeneous Graph}
When training GNNs in a supervised learning manner, a large amount of labeled data is required. For example, for different tasks working on the same graph, we need multiple labeled datasets to train GNNs. However, accessing sufficient labeled datasets for each task is sometimes impossible. To reduce the reliance on labeled data, pre-training strategies are broadly adopted. The goal of the pre-training is to learn generic initialization parameters for the model by using easily accessible unlabeled data.

The pre-training strategy typically involves two steps. First, pre-training a model $f_{\boldsymbol{\Theta}}$ on large-scale unlabeled data. Then, the pre-trained model is fine-tuned on specific downstream tasks. Given a heterogeneous graph $\mathcal{G}$ split into $\mathcal{G}_{pre}$ and $\mathcal{G}_{fine}$, where $\mathcal{G}_{pre}$ is the graph for pre-training and $\mathcal{G}_{fine}$ is the graph for fine-tuning. In the first step (pre-train a model), the objective to optimize is:
\begin{equation}
    \label{ssy1210:obj_pretrain}
    \boldsymbol{\Theta}_{pre} = \mathop{\arg\min}\limits_{\boldsymbol\Theta}\mathcal{L}_{pre}\left(f_{\boldsymbol{\Theta}};\mathcal{G}_{pre}\right),
\end{equation}
where $\mathcal{L}_{pre}$ and $\boldsymbol{\Theta}_{pre}$ represent the loss function and optimized parameter in the well-designed pre-training task, respectively. The parameter $\boldsymbol{\Theta}_{pre}$ is expected to capture  transferable knowledge in the heterogeneous graph (e.g., the intrinsic structure of the graph), which tends to provide a good parameter initialization for the model in downstream tasks. In the second step (fine-tune a model), it is required to optimize the loss function in the downstream task:
\begin{equation}
    \label{ssy1210:obj_finetine}
    \begin{aligned}
    &\mathop{\min}\limits_{\boldsymbol\Theta}\mathcal{L}_{fine}\left(f_{\boldsymbol{\Theta}};\mathcal{G}_{fine}\right), \\
        &\mathrm{s.t.} \ \boldsymbol{\Theta}_{init}=\boldsymbol{\Theta}_{pre},
    \end{aligned}
\end{equation}
where $\mathcal{L}_{fine}$ is the loss function in the fine-tuning process. The constraint in Eq.~\ref{ssy1210:obj_finetine} implies that the model in the fine-tuning is initialized by the pre-trained parameter $\boldsymbol{\Theta}_{pre}$.

\textcolor{black}{\subsection{Transformer-based Heterogeneous Encoding}
Inspired by the success of the Transformer~\cite{DBLP:conf/www/HuDong20,DBLP:journals/JEAI/LiuQin25}, we adopt transformer-based heterogeneous encoding to provide node representations for subsequent pre-training processes. Formally, the update of node $v$'s representation can be formulated as:
\begin{equation}
    \label{ssy1210:node_update}
    \left\{
    \begin{aligned}
        &\mathbf{h}_{v}^{(0)}=\mathbf{x}_{v}^{(0)},\\
        &\mathbf{h}_{v}^{(i)}=\sum_{u \in \mathcal{N}_{v}}{\rm{softmax}}_{u \in \mathcal{N}_{v}}\left(\mathop{\parallel}_{j=1...h^{\prime}}\frac{\mathbf{k}_{u}^{j}\mathbf{W}^{\prime}_{f_\mathcal{E}(e_{uv})}{\mathbf{q}_{v}^{j}}^{\top}}{\sqrt{d}}\right)\odot\mathop{\parallel}_{j=1...h^{\prime}}\mathbf{m}_{u}^{j}\mathbf{W}^{\prime\prime}_{f_\mathcal{E}(e_{uv})},~~~i=1,...,K,
    \end{aligned}
    \right.
\end{equation}
where $\mathbf{x}_{v}^{(0)}$ is the initial feature of node $u$, $\mathbf{h}_{v}^{(i)}$ is the representation of node $u$ at the $i$-th layer, $\mathbf{W}^{\prime}_{f_\mathcal{E}(e_{uv})}$ and $\mathbf{W}^{\prime\prime}_{f_\mathcal{E}(e_{uv})}$ are two trainable matrices related to the edge type $f_\mathcal{E}(e_{uv})$, $h^{\prime}$ is the number of heads, and
\begin{equation}
    \label{ssy1210:QKV}
    \mathbf{k}_{u}^{j} = K^{j}_{f_{\mathcal{V}}(u)}\left(\mathbf{h}^{(i-1)}_{u}\right), \mathbf{q}_{v}^{j} = Q^{j}_{f_{\mathcal{V}}(v)}\left(\mathbf{h}_{v}^{(i-1)}\right),\mathbf{m}_{u}^{j} = M^{j}_{f_{\mathcal{V}}(u)}\left(\mathbf{h}^{(i-1)}_{u}\right),
\end{equation}
\textcolor{black}{are output vectors of the learnable linear mapping functions $K^{j}_{f_{\mathcal{V}}(u)}\left(\cdot\right)$, $Q^{j}_{f_{\mathcal{V}}(u)}\left(\cdot\right)$, $M^{j}_{f_{\mathcal{V}}(u)}\left(\cdot\right)$, respectively. $\mathbf{k}_{u}^{j}$, $\mathbf{q}_{u}^{j}$, and $\mathbf{m}_{u}^{j}$ correspond to the \textit{Key} vector, \textit{Query} vector, and \textit{Value} vector in the classic Transformer architecture in that order. Eq. \ref{ssy1210:node_update} essentially follows the traditional Transformer computation, where the product of the \textit{Key} and \textit{Query} vectors is first softmaxed and then multiplied with the \textit{Value} vector. It is important to note that, unlike the naive Transformer, which directly computes the dot product between \textit{Key}, \textit{Query}, and \textit{Value}, the Graph Transformer in Eq. \ref{ssy1210:node_update} introduces distinct edge-based learnable matrices $\mathbf{W}^{\prime}_{f_\mathcal{E}(e_{uv})}$ and $\mathbf{W}^{\prime\prime}_{f_\mathcal{E}(e_{uv})}$ to capture the rich semantics in heterogeneous graphs.}}

\subsection{Structure-Aware Pre-training Task}
\label{ssy1210:task1_query}
In contrastive learning, a typical triplet structure is \textless query
sample, positive sample, negative sample\textgreater. The goal of contrastive learning is to learn representations so that the query sample is close to the positive sample and far away from the negative sample. Based on this triplet structure, we design pre-training tasks with enhancements. In this section, we propose the structure-aware pre-training task with enhanced query samples, which aims to capture fine-grained heterogeneous information.

Given a query sample ${\mathbf{{h}}}_{v}\in\mathbb{R}^{d}$ (obtained from the Transformer-based heterogeneous encoding), a straightforward idea is to directly use the query sample ${\mathbf{{h}}}_{v}$ without any enhancement. However, directly using the query sample ${\mathbf{{h}}}_{v}$ makes GNNs only focus on node-level information in the graph, which leads to the ignorance of higher-order properties in heterogeneous graphs. In order to take full advantage of the rich structures in the heterogeneous graph, the network-schema subspace $\mathbf{{S}}_{1}$ is constructed to help enhance ${\mathbf{{h}}}_{v}$. Specifically, we choose the embeddings of neighboring nodes of the node $v$ to be the columns of the subspace for node $v$, i.e.,
\begin{equation}
    \label{ssy1210:network_subspace}
    \mathbf{{S}}_{1}^{v}=\left[\mathbf{{h}}_{u_1}^{\varpi_1},\cdots,\mathbf{{h}}_{u_k}^{\varpi_1},\mathbf{{h}}_{u_{k+1}}^{\varpi_2},\cdots,\mathbf{{h}}_{u_{k+m}}^{\varpi_2},\cdots,\mathbf{{h}}_{u_{l}}^{\varpi_n}\right],
\end{equation}
where $\left\{u_i\right\}_{i=1}^{l}$ are neighbor nodes of the node $v$ and $\left\{\varpi_i\right\}_{i=1}^{n}$ are types of neighbor nodes (e.g., $f_{\mathcal{V}}(v_1)=\varpi_1$). Using the constructed subspace, the enhanced query sample ${\mathbf{{h}}}^{\prime}_{v}$ can be written as:
\begin{equation}
    \label{ssy1210:enhanced_query_sample}
    {\mathbf{{h}}}^{\prime}_{v}=\mathbf{{S}}_{1}^{v}\mathbf{{p}}^{v},
\end{equation}
where $\mathbf{{p}}^{v}$ is the  importance vector. In order to preserve the fine-grained heterogeneity in the graph, we decompose the importance vector $\mathbf{{p}}^{v}$ into two parts, i.e.,
\begin{equation}
    \label{ssy1210:enhance_querysample}
    {\mathbf{{h}}}^{\prime}_{v}={{\mathbf{{S}}}_{1}^{v}}\times({\mathbf{{p}}^{v}_{a}\odot{\mathbf{{p}}^{v}_{b}}}),
\end{equation}
where $\mathbf{{p}}^{v}_{a}$ assigns weights to different nodes with the same type and $\mathbf{{p}}^{v}_{b}$ assigns weights to different node types. The element-wise product $\odot$ illustrates that the calculation of weight is a two-stage process and the final weight is subject to both the type-level weight and the node-level weight. In this part, the attention mechanism is used to estimate the two types of importance vectors mentioned above. The node-level weight $\mathbf{{p}}^{v}_{a}$ can be written as:
\begin{equation}
\label{ssy1210:node_attention}
\mathbf{{p}}^{v}_{a}=\left[{\alpha_{v,u_1},\cdots,\alpha_{v,u_k}},\alpha_{v,u_{k+1}},\cdots,\alpha_{v,u_{k+m}},\cdots,\alpha_{v,u_{l}}\right]^{\top},
\end{equation}
where $\alpha_{v,u_{i}}$ is the contribution of the node $u_{i}$ to node $v$ within node type $f_{\mathcal{V}}(u_{i})$. Specifically, $\alpha_{v,u_{i}}$ can be written as follows:
\begin{equation}
    \label{ssy1210:contri_same_type}
    \alpha_{v,u_{i}}=\frac{\exp \left(\text {LeakyReLU }\left({\mathbf{{a}}}_{\varpi_{k}}^{\top}\left[\mathbf{{h}}_{v} \| \mathbf{{h}}_{u_{i}}\right]\right)\right)}{\sum_{u_j \in \mathcal{N}_{v}^{\varpi_{k}}} \exp \left(\operatorname{LeakyReLU}\left(\mathbf{{a}}_{\varpi_{k}}^{\top}\left[ \mathbf{{h}}_{v} \| \mathbf{{h}}_{u_j}\right]\right)\right)},
\end{equation}
\textcolor{black}{where $\mathbf{{a}}_{\varpi_{i}}\in\mathbb{R}^{2d\times1}$ is the learnable attention vector for node $v^{\prime}s$ neighbor nodes with node type ${\varpi_{i}}$. The existence of $\mathbf{{a}}_{\varpi_{i}}$ is to integrate diverse node-type information within heterogeneous graphs \cite{DBLP:conf/www/WangJi21,DBLP:journals/tois/YangHu21}.}

Similarly, we resort to the attention mechanism to obtain the type weight $\mathbf{{p}}^{v}_{b}$. It should be noted that the calculation of the type-level weight is different from the calculation of the node-level weight. This is because the type weight is calculated in a global perspective, which means that the type-level weight should be an average of the type-level weights of all nodes rather than determined by the type-level weight of a single node. Using the subspace in Eq.~\ref{ssy1210:network_subspace} and the $\mathbf{{p}}^{v}_{a}$ in Eq.~\ref{ssy1210:node_attention}, the type-level embedding can be written as follows:
\begin{equation}
    \label{ssy1210:type_embedding}
    \mathbf{{h}}_{\varpi_{k}}^{v} = \sigma\left(\sum_{u_i \in \mathcal{N}_{v}^{\varpi_{k}}}\alpha_{v,u_{i}}\mathbf{{h}}_{u_i}^{\varpi_k}\right), k=1,\cdots,n,
\end{equation}
where $\mathbf{{h}}_{\varpi_{k}}^{v}$ is the fused embedding for node type ${\varpi_{k}}$. After getting fused embeddings, we then need to calculate the type weight. The type weight $\mathbf{{p}}^{v}_{b}$ can be presented as:
\begin{equation}
\label{ssy1210:type_attention}
\begin{aligned}
\mathbf{{p}}^{v}_{b}=
[{\beta_{v,\varpi_{1}},\cdots,\beta_{v,\varpi_{1}},}{\beta_{v,\varpi_{2}},\cdots,\beta_{v,\varpi_{2}},} \cdots,\beta_{v,\varpi_{n}}]^{\top},
\end{aligned}
\end{equation}
where $\beta_{v,\varpi_{k}}$ is the contribution of the node type $\varpi_{k}$ to node $v$. Specifically, $\beta_{v,\varpi_{k}}$ can be calculated as:
\begin{equation}
    \label{ssy1210:contri_diff_type}
    \beta_{v,\varpi_{k}}=\frac{\exp\left(\frac{1}{|\mathcal{V}_{q}|} \sum_{v \in \mathcal{V}_{q}} \mathbf{ a}_{t}^{\top} \cdot \tanh \left(\mathbf{{W}}_{t} \mathbf{{h}}_{\varpi_{k}}^{v}+\mathbf{{b}}_{t}\right)\right)}{\sum_{i=1}^n \exp \left(\frac{1}{|\mathcal{V}_{q}|} \sum_{v \in \mathcal{V}_{q}} \mathbf{ a}_{t}^{\top} \cdot \tanh \left(\mathbf{{W}}_{t} \mathbf{{h}}_{\varpi_{i}}^{v}+\mathbf{{b}}_{t}\right)\right)},
\end{equation}
where $\mathbf{ a}_{t}$ is the attention vector for node types, $\mathbf{{W}}_{t}$ and $\mathbf{ b}_{t}$ are trainable parameters that transform fused type-level embeddings into the same space, and $\mathcal{V}_q$ is the set of nodes corresponding to the same type of query samples ({i.e., the type of nodes in $\mathcal{V}_{q}$ in Eq.~\ref{ssy1210:contri_diff_type} is the same as that of node $v$}). Plugging Eq.~\ref{ssy1210:network_subspace}, Eq.~\ref{ssy1210:node_attention} and Eq.~\ref{ssy1210:type_attention} into Eq.~\ref{ssy1210:enhance_querysample}, we can obtain the final enhanced query sample ${\mathbf{{h}}}^{\prime}_{v}$.

\begin{figure}[h]
    \centering
    \includegraphics[width=1\linewidth]{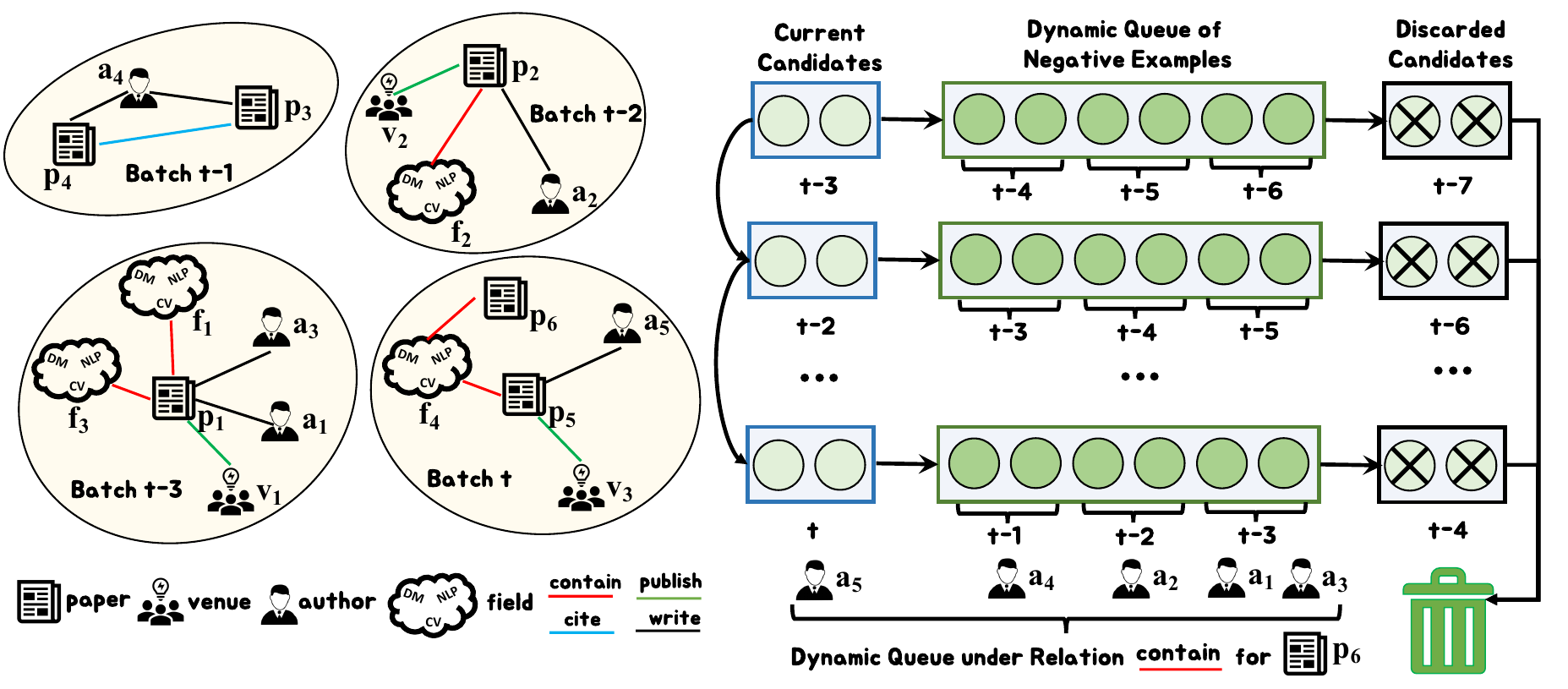}
    \caption{{Display of the dynamic negative sample queue. Only the most recent batch of negative samples are kept in the queue for use in the current batch.}}
    \label{ssy1210:queue_framework}
\end{figure}

In order to conduct contrastive learning, the negative sampling is indispensable. Most of the traditional negative sampling strategies are designed for homogeneous graphs, that is, selecting unconnected nodes of the query node as negative samples. However, heterogeneous graphs are distinct from homogeneous graphs. Specifically, heterogeneous graphs contain many different types of nodes and edges, which makes simply selecting unconnected nodes as negative samples cause semantic misleading. For example, given a positive triplet $\left(paper_i, citing, paper_j\right)$, a reasonable negative sample is the $paper_k$ that is not cited by the $paper_i$, rather than the $field_q$ that is not connected to the $paper_i$. To tackle the above problem, two negative sampling strategies are designed in this part. Given a positive triplet $\left(v,r,u\right)$, we first perform negative sampling based on the specific relationship $r$ between node $v$ and node $u$, i.e.,
\begin{equation}
    \label{ssy1210:neg_sample_part1_query}
    \mathcal{S}_{\left( v, r, u\right)_{t}}^{1}=\left\{u^{-} \mid {\emph A}_{v u^{-}}^{r}=0, u^{-} \in \mathcal{V}_{t}, f_{\mathcal{V}}(u)=f_{\mathcal{V}}(u^{-})\right\},
\end{equation}
where $\mathcal{S}_{\left( v, r, u\right)_{t}}^{1}$ is a set of relationship-based negative sampling nodes for the structure-aware pre-training task, $t$ is the current time (i.e., the current batch), $u^{-}$ represents a negative sample in $\mathcal{S}_{\left( v, r, u\right)_{t}}^{1}$, $\mathcal{V}_{t}$ is the node set of the current batch ({the subscript $t$ is used to indicate different batches}), and ${\mathbf A}^{r}$ is the adjacency matrix under the relation $r$. To further improve the quality of negative samples, we then introduce a dynamic queue {(as shown in Fig. \ref{ssy1210:queue_framework})} to prevent self-supervised learning from only focusing on the information contained in the current batch. Negative samples from a dynamic queue can be written as:
\begin{equation}
    \label{ssy1210:neg_sample_part2_query}
    \mathcal{S}^{1}_{Queue}=\mathcal{S}_{\left( v, r, u\right)_{t-1}}^{1}\cup\cdots\cup\mathcal{S}_{\left( v, r, u\right)_{t-M}}^{1},
\end{equation}
where $\mathcal{S}^{1}_{Queue}$ is the union of negative sample sets in the latest $M$ batches (from time $t-M$ to $t-1$). To sum up, the final set $\mathcal{S}^{(1)}$ of negative samples for the enhanced query sample can be written as:
\begin{equation}
\label{ssy1210:neg_sample_query_final}
    \mathcal{S}^{(1)}=\bigcup_{i=t-M}^{t} \mathcal{S}_{\left( v, r, u\right)_{i}}^{1}.
\end{equation}
 After obtaining the enhanced query sample and negative samples $ \mathcal{S}^{(1)}$, the contrastive loss we need to optimize can be written as:
 \begin{equation}
     \label{ssy1210:loss1}
     \mathcal{L}_{1}=-\mathbb{E}_{{\mathcal{P}_{v}}}\left[\log \frac{\exp \left({\left[{{\mathbf{{S}}}_{1}^{v}}\times({\mathbf{{p}}^{v}_{a}\odot{\mathbf{{p}}^{v}_{b}}})\right]^{\top} \mathbf{ W}_{r} \mathbf{ h}_{u}} {\tau}\right)}{\sum_{j \in\{u\} \cup \mathcal{S}^{(1)}} \exp \left(\left[{{\mathbf{{S}}}_{1}^{v}}\times({\mathbf{{p}}^{v}_{a}\odot{\mathbf{{p}}^{v}_{b}}})\right]^{\top} \mathbf{ W}_{r} \mathbf{ h}_{j} / \tau\right)}\right],
 \end{equation}
 where $\mathcal{P}_{v}$ is the set of all positive triples, $\mathbf{{W}}_r$ is a trainable matrix for the specific relationship $r$, and $\tau$ is a temperature coefficient.
\subsection{Semantic-Aware Pre-training Task}
To solve the problem of semantic mismatch mentioned in Section \ref{ssy1210:introduction}, a semantic-aware pre-training task with enhanced positive/negative samples is designed. As mentioned in Section \ref{ssy1210:introduction}, the key to solving the problem of semantic mismatch is to incorporate semantic neighbors. An effective method is proposed in this section to capture information of semantic neighbors by constructing a perturbation subspace, where each column of the perturbation subspace is the output of a perturbed GNN. When the magnitude of the perturbation is not large, each column in the perturbation subspace can be regarded as an embedding of semantic neighbors~\cite{DBLP:conf/www/XiaWu22,DBLP:conf/ACL/WuWu22}. Essentially, this is because GNNs with similar parameters have similar outputs, which ensures that they have similar semantics.

Denoting a positive/negative sample obtained from a GNN encoder $f_{\boldsymbol\Theta}$ as ${\mathbf{{h}}}_{u}=f_{\boldsymbol\Theta}\left(u,\mathcal{G}\right)$, then the perturbation subspace $\mathbf{{S}}_{2}^{u}$ designed for the mismatched situation can be written as:
\begin{equation}
    \label{ssy1210:perturb_subspace}
    \mathbf{{S}}_{2}^{u} = \left[\mathbf{{h}}_{u^{\prime}_{1}},\mathbf{{h}}_{u^{\prime}_{2}},\cdots,\mathbf{{h}}_{u^{\prime}_{q}}\right],
\end{equation}
where $\{\mathbf{{h}}_{u^{\prime}_{i}}\}^{q}_{i=1}$ are the outputs of perturbed GNNs, i.e.,
\begin{equation}
    \label{ssy1210:perturb_gnn}
    \mathbf{{h}}_{u^{\prime}_{i}} = f^{perturb}_{\boldsymbol\Theta}\left(u,\mathcal{G}\right).
\end{equation}
Defining the parameters in the original GNN as $\boldsymbol\Theta=\left[\boldsymbol{\Theta}_{1},\cdots,\boldsymbol{\Theta}_{k}\right]$, then the method used to perturb the GNN $f_{\boldsymbol{\Theta}}$ can be written as:
\begin{equation}
    \label{ssy1210:perturb_para}
    \boldsymbol{\Theta}_{i}^{perturb} = \boldsymbol{\Theta}_{i} + U(-\mu,\mu)\times std(\boldsymbol{\Theta}_{i}),\quad \quad i=1,\cdots,k,
\end{equation}
where $std$ represents the standard deviation, $\mu$ is a hyper-parameter, and $U(-\mu,\mu)$ is the uniform distribution with noise ranged from $-\mu$ to $\mu$. When $\mu$ takes a small value, each column in the subspace can be regarded as the embedding of the semantic neighbors of node $u$. Essentially, this is because similar parameter tensors make the perturbed GNN and the original GNN have similar outputs. Using the constructed perturbation subspace, the enhanced positive/negative sample can be written as:
\begin{equation}
    \label{ssy1210:enhanced_positive_negative_sample}
    {\mathbf{{h}}}^{\prime}_{u}=\mathbf{{S}}_{2}^{u}\mathbf{{p}}^{u},
\end{equation}
where $\mathbf{{p}}^{u}$ is the corresponding importance vector. Considering that the noise used to perturb the GNN here is of the same type (uniform distribution), we assign the same weight to each column of the perturbation subspace, which means that each generated semantic neighbor has the same contribution on the enhancement of the positive/negative sample, i.e., $\mathbf{p}^{u}=\left[\frac{1}{q},\cdots,\frac{1}{q}\right]$. {The Eq.~\ref{ssy1210:enhanced_positive_negative_sample} can be regarded as a linear combination of the columns of the perturbation subspace. The reasons why ``linear combination'' can help deal with mismatched situations are as follows: i) Intuitively, since each column of the perturbation subspace can be regarded as the semantic neighbor of positive samples/negative samples, the linear combination of the columns of the perturbation subspace aggregates the information of semantic neighbors. Therefore, the embedding after the linear combination can be regarded as a prototype of all semantic neighbors to a certain extent, which contains the common features of semantic neighbors and thus helps the model to learn the knowledge with better transferability (general knowledge). Similar ideas can be seen in \cite{DBLP:conf/nips/CaronMisra20}.} ii) Mathematically, the so-called semantic mismatch refers to the difference $\mathbf{\epsilon}\in\mathbb{R}^{d}$ between the original data $\mathbf{y}^{o}\in\mathbb{R}^{d}$ and the ideal data $\mathbf{y}^{i}\in\mathbb{R}^{d}$ (i.e., $\mathbf{y}^{i}$ = $\mathbf{y}^{o}$ + $\mathbf{\epsilon}$). The perturbation subspace used here is expected to estimate this difference $\epsilon$. In other words, using the perturbation subspace $\mathbf{S}=\left[\mathbf{y}^{o^\prime}_1,\cdots, \mathbf{y}^{o^\prime}_q\right]\in\mathbb{R}^{d\times q}$ (where $\mathbf{y}^{o^\prime}_l|^{q}_{l=1}$ is a perturbed version of $\mathbf{y}^{o}$) gives the model the ability to describe the difference $\mathbf{\epsilon}$. Specifically, without using a subspace (using $\mathbf{y}^{o}$ directly, $\mathbf{\hat y}^{i}=\mathbf{y}^{o}$), $\mathbf{\hat y}^{i}$ can only represent a point in $d$-dimensional space. The use of subspace $\mathbf{S}$ allows the model to use a $q$-dimensional hyperplane (linear combination of columns of the matrix $\mathbf{S}$) in $d$-dimensional space to describe $\epsilon$ (i.e., $\mathbf{\hat y}^{i}=\mathbf{S}_{{d\times q}}\mathbf{p}_{{q\times1}}$). Compared with a point in $d$-dimensional space, the $q$-dimensional hyperplane obviously has a better ability to describe the difference.

Similar to Section \ref{ssy1210:task1_query}, we also design two negative sampling strategies. Specifically, given a positive triplet $\left(v,r,u\right)$ (for example, $(paper_i, citing, paper_j)$), we first conduct negative sampling based on the relation $r$, and then we perform negative sampling by using a dynamic queue to ensure that the model does not only focus on the current batch. Therefore, the final set of negative samples $\mathcal{S}^{(2)}$ in this part can be written as:
\begin{equation}
    \label{ssy1210:neg_sample_positive_enhanced}
    \mathcal{S}^{(2)}=\mathcal{S}_{\left( v, r, u\right)_{t}}^{2}\cup\mathcal{S}^{2}_{Queue} \,,
\end{equation} 
where $\mathcal{S}_{\left( v, r, u\right)_{t}}^{2}$ is a set of enhanced negative samples under the current batch and $\mathcal{S}^{2}_{Queue}$ is a set of negative samples collected from the dynamic queue. Noting that the $\mathcal{S}^{2}_{Queue}$ stores the enhanced negative samples from the latest batches, which is different from the $\mathcal{S}^{1}_{Queue}$ that stores the original negative samples.

After obtaining the enhanced positive and negative samples, the contrastive loss we need to optimize can be written as follows:
 \begin{equation}
     \label{ssy1210:loss2}
     \mathcal{L}_{2}=-\mathbb{E}_{{\mathcal{P}_{v}}}\left[\log \frac{\exp \left(\mathbf{{h}}_{v}^{\top} \mathbf{{W}}_{r} \mathbf{ S}_{2}^{u}\mathbf{{p}}^{u} / \tau\right)}{\sum_{j \in\{u\} \cup \mathcal{S}^{(2)}} \exp \left(\mathbf{{h}}_{v}^{\top} \mathbf{{W}}_{r} \mathbf{ S}_{2}^{j}\mathbf{{p}}^{j} / \tau\right)}\right].
 \end{equation}

\textcolor{black}{\subsection{Large-Scale Heterogeneous Graph Pre-training}
\label{ssy0206:batch_training}
Heterogeneous graphs in the real world are usually large-scale (containing millions of nodes). However, extending pre-training strategies to large-scale graphs is not straightforward. For example, the clustering-based pre-training method requires label propagation over the entire graph to obtain high-quality pseudo-labels for guiding the pre-training process, while performing label propagation on a large-scale graph with millions of nodes is extremely challenging \cite{DBLP:conf/nips/YangGuan22}.}

\textcolor{black}{Instead, we focus on pre-training on large-scale heterogeneous graphs, which is more closely aligned with real-world scenarios. The heterogeneous graphs used in this paper are too large to be directly used for the pre-training process. Inspired by the HGSampling~\cite{DBLP:conf/www/HuDong20}, we adapt the proposed method to operate in a mini-batch manner, fitting it into the hardware constraints. More specifically, we maintain a bucket for each type of node in the heterogeneous graph, and employ an importance sampling strategy to sample the same number of nodes for each type. Given a sampled node $v$, we add $v$'s all direct neighbors into the corresponding budget and $v$'s normalized degree to these neighbors. It should be noted that the normalized degree is proportional to the sampling probability, i.e., the nodes with higher values in the corresponding budget have a higher probability of being sampled. The overall sampling process can be summarized as follows: Given the node type $\tau$, we first sample $n$ nodes of this type based on the sampling probability. These newly sampled nodes are added to the output node set and their neighborhoods are updated in the corresponding budget. Then, we repeat this procedure $L$ times and obtain a sampled sub-graph with a depth of $L$ from the initial nodes.} The overall framework of the proposed PHE is depicted in Fig. \ref{ssy1210:framework}. We jointly optimize the above two pre-training tasks and the objective of the proposed PHE can be written as the sum of the losses:
\begin{equation}
    \label{ssy1210:loss}
    \mathcal{L} = \mathcal{L}_{1} + \lambda\mathcal{L}_{2},
\end{equation}
where $\lambda$ is a hyper-parameter to balance the effect of two pre-training tasks. The detailed algorithm of the proposed method is provided in Algorithm 1.\\
\begin{algorithm}[h]
    \caption{The training algorithm of PHE}
    \begin{algorithmic}[1] 
        \REQUIRE A heterogeneous graph $\mathcal{G}$, a GNN $f_{\boldsymbol{\Theta}}$ 
        \ENSURE The pre-trained parameters $\boldsymbol{\Theta}_{pre}$
        \FOR {each sampled batch $\mathcal{V}$ in $\mathcal{G}$}
            \FOR {each positive triplet $(v,r,u)$ in $\mathcal{V}$}
                \STATE{Construct the network-schema subspace $\mathbf{{S}}_{1}^{v}$ by \eqref{ssy1210:network_subspace}}
                \STATE{Compute fine-grained weights $\alpha_{v,u},\beta_{v,\varpi}$ by \eqref{ssy1210:contri_same_type} and \eqref{ssy1210:contri_diff_type}}
                \STATE{Prepare  the enhanced query sample $\mathbf{{h}}_{v}^{\prime}$ by \eqref{ssy1210:enhance_querysample}}
                \STATE{Prepare the negative samples $\mathcal{S}^{(1)}$ with \eqref{ssy1210:neg_sample_query_final}}
                \STATE{Calculate $\mathcal{L}_{1}$ by \eqref{ssy1210:loss1}}
                \STATE{Construct the perturbation  subspace $\mathbf{{S}}_{2}^{u}$ by \eqref{ssy1210:perturb_subspace}}
                \STATE{Prepare  the enhanced positive sample $\mathbf{{h}}_{u}^{\prime}$ by \eqref{ssy1210:enhanced_positive_negative_sample}}
                \FOR {each negative sample $u^{-}$ in $\mathcal{S}_{v}$}
                    \STATE {Construct the perturbation  subspace $\mathbf{{S}}_{2}^{u^{-}}$ by \eqref{ssy1210:perturb_subspace}}\\
                    \STATE {Prepare  the enhanced negative sample $\mathbf{{h}}_{u^{-}}^{\prime}$ by \eqref{ssy1210:enhanced_positive_negative_sample}}\\
                \ENDFOR
                \STATE {Prepare the negative samples $\mathcal{S}^{(2)}$ with \eqref{ssy1210:neg_sample_positive_enhanced}}\\
                \STATE {Calculate $\mathcal{L}_{2}$ by \eqref{ssy1210:loss2}}\\
            \ENDFOR\\
            \STATE {Calculate total loss $\mathcal{L}$ by \eqref{ssy1210:loss}}\\
            \STATE {Update the trainable parameters $\boldsymbol{\Theta}$}
        \ENDFOR
    \end{algorithmic} 
\end{algorithm}

\textcolor{black}{\subsection{Model Complexity Analysis}
\label{ssy1210:complexity}
In this section, we provide the detailed model complexity analysis for the proposed PHE. Assuming the node feature dimension is $d$, and the model training  contains $n_{\mathcal{B}}$ batches in total (as shown in Section \ref{ssy0206:batch_training}). The complexity of the forward process for node encoding is $\mathcal{O}(\sum\limits_{i=1}^{n_{\mathcal{B}}}cX^{i}_{\rm {gnn}})$, where $c$ the number of perturbations and $X^{i}_{\rm {gnn}}$ is the time complexity of message passing and aggregation in a specific GNN. The time complexity of calculating the fine-grained contribution in structure-aware pre-training task is $O(\sum\limits_{i=1}^{n_{\mathcal{B}}}\vert\mathcal{V}^{i}_{\mathcal{B}}\vert(\bar{n}_{\mathcal{N}}+n_{\mathcal{T}})d^2)$, where $\mathcal{V}^{i}_{\mathcal{B}}$ is the node set in $i$-th batch, $\bar{n}_{\mathcal{N}}$ is the average number of neighbor nodes of a node in $\mathcal{V}^{i}_{\mathcal{B}}$, and $n_{\mathcal{T}}$ is the number of node types. In contrastive optimization, the time complexity of calculating the similarity between query samples and the positive/negative samples can be written as $\mathcal{O}(\sum\limits_{i=1}^{n_{\mathcal{B}}}\vert\mathcal{V}^{i}_{\mathcal{B}}\vert(\vert\mathcal{P}_{\rm{Sem}}^{+}\vert+\vert\mathcal{P}_{\rm{Sem}}^{-}\vert+\vert\mathcal{P}_{\rm{Str}}^{+}\vert+\vert\mathcal{P}_{\rm{Str}}^{-}\vert)d^2)$, where $\mathcal{P}_{\rm{Sem}}^{+}$/$\mathcal{P}_{\rm{Sem}}^{-}$ and $\mathcal{P}_{\rm{Str}}^{+}$/$\mathcal{P}_{\rm{Str}}^{-}$ are the positive/negative sample set in structure-aware and semantic-aware pre-training task, respectively. Therefore, the overall complexity of the proposed method is $\mathcal{O}(\sum\limits_{i=1}^{n_{\mathcal{B}}}cX^{i}_{\rm {gnn}}+\vert\mathcal{V}^{i}_{\mathcal{B}}\vert(\bar{n}_{\mathcal{N}}+n_{\mathcal{T}})d^2+\vert\mathcal{V}^{i}_{\mathcal{B}}\vert(\vert\mathcal{P}_{\rm{Sem}}^{+}\vert+\vert\mathcal{P}_{\rm{Sem}}^{-}\vert+\vert\mathcal{P}_{\rm{Str}}^{+}\vert+\vert\mathcal{P}_{\rm{Str}}^{-}\vert)d^2)$. Given that the constant $c$ related to the perturbations is relatively small (e.g., $c\in[3,5]$) and the computation of fine-grained contributions remains on the order of $\mathcal{O}(d^2)$, the time complexity of the proposed method is of the same order as that of existing methods (we will verify this in subsequent experiments). }\\

\textbf{Discussion.} {\textcolor{black}{The proposed methods can be flexibly extended to other architectures by appropriately modifying the heterogeneous encoding module. This property allows the proposed method to have a wider range of applications. We will discuss the flexible architecture generalization of the proposed method more thoroughly in the subsequent experimental section.} In addition, we only focus on one-sided sample enhancement (query sample side or positive/negative sample side) in each pre-training task. The reason for the one-sided sample enhancement is twofold: First, one-sided enhancement ensures that we use at least a part of the original graph data (i.e., samples from the side without enhancement), which is beneficial to preserve the original information in the data. Second, different sides have different emphases. In other words, we pay more attention to the capture of high-order heterogeneous structural properties on the pre-training task with enhanced query samples, while on the positive/negative sample side enhancement, we focus more on how to deal with mismatched situations. {Furthermore, to alleviate the system’s
management/storage burdens associated with negative samples, we set a maximum length for the negative sample queue used in structure-aware and semantic-aware tasks. Specifically, we maintain the queue length by dynamically updating it—adding the latest negative samples while removing the oldest ones. More details about the negative sample queue can be seen in Fig. \ref{ssy1210:queue_framework} and Section \ref{ssy1210:implement_details}. Essentially, the proposed pre-training strategies increase the diversity of contrastive views by integrating heterogeneous information (semantic and structural information) into the construction of contrastive views. Recently, several methods~\cite{DBLP:conf/www/XiaWu22,DBLP:journals/tkde/YuXia22,DBLP:journals/tkdd/LiuHao24} have found that adding noise (in the input space or latent space) is beneficial for constructing diverse views, which in turn helps improve model performance. The theoretical foundation for this performance gain lies in the improvement of feature learning via noise injection. Tamkin \emph{et al.} \cite{DBLP:conf/nips/TamkinGlasgow23} have proved that adding noise to contrastive views enables the learned representations to better align with the direction of optimal representations, thereby improving model performance (Theorem 4.1 in \cite{DBLP:conf/nips/TamkinGlasgow23}).}

\section{Experiments}
\label{ssy1210:experiment}
In this section, we assess the performance of the proposed method by conducting extensive experiments. Our experiments aim to answer the following research questions (RQs):
\begin{itemize}[leftmargin=*]
\item[$\bullet$]  \textbf{RQ1:} Compared with the existing methods, can the proposed PHE achieve performance improvements in heterogeneous graphs? 
\item[$\bullet$] \textbf{RQ2:} What is the ability of the proposed PHE to learn transferable knowledge in heterogeneous graphs? 
\item[$\bullet$] \textbf{RQ3:} For different pre-training strategies in the proposed PHE, what are their impacts on performance?
\end{itemize}
\subsection{Experiment Settings}
\subsubsection{\textbf{Datasets}} { We conduct experiments on the large-scale heterogeneous graph---Open Academic Graph (OAG)~\cite{DBLP:conf/kdd/ZhangLiu19}
. The OAG is a billion-scale graph that unifies the two academic graphs Microsoft Academic Graph and AMiner. To verify whether the proposed pre-training strategies can learn transferable knowledge, we select three subgraphs of different domains (i.e., the computer science domain, the engineering domain, and the material domain) in the OAG to conduct experiments. 

The statistics of the real-world datasets are summarized in Table \ref{ssy1210:dataset_detail}. As shown in Table \ref{ssy1210:dataset_detail}, each dataset contains multiple types of nodes and edges. In the experiments, we divide each dataset into training, validation, and test sets according to the ratio of 8:1:1.
\begin{table}[h]
\renewcommand\arraystretch{1.2}
  \centering
  \caption{Statistics of datasets used in experiments.}
  \scalebox{0.9}{
    \begin{tabular}{ccccc}
    \toprule[1.5pt]
    \textbf{Dataset} & \textbf{Node Type} & \textbf{\#Nodes} & \textbf{Edge Type} & \textbf{\#Edges} \\
    \midrule
    \multicolumn{1}{c}{\multirow{4}[2]{*}{\makecell[c]{Computer\\Science}}} & venue & 27,433 & paper-venue & 5,597,606 \\
          & author & 5,985,759 & paper-author & 15,571,614 \\
          & field & 289,930 & paper-field & 47,462,559 \\
          & paper & 5,597,605 & paper-paper & 31,441,552 \\
    \midrule
    \multicolumn{1}{c}{\multirow{4}[2]{*}{Engineering}} & venue & 19,867 & paper-venue & 3,239,504 \\
          & author & 1,819,100 & paper-author & 3,741,135 \\
          & field & 99,444 & paper-field & 22,498,822 \\
          & paper & 3,239,504 & paper-paper & 4,848,158 \\
    \midrule
    \multicolumn{1}{c}{\multirow{4}[2]{*}{Materials}} & venue & 15,141 & paper-venue & 2,442,235 \\
          & author & 2,005,362 & paper-author & 5,582,765 \\
          & field & 79,305 & paper-field & 19,119,947 \\
          & paper & 2,442,235 & paper-paper & 13,011,272 \\
    \bottomrule[1.5pt]
    \end{tabular}}%
  \label{ssy1210:dataset_detail}%
\end{table}}
\subsubsection{\textbf{Pre-training and Fine-tuning}}{The goal of pre-training is to learn general initialization parameters for models in the downstream tasks by utilizing a large amount of unlabeled data, which can make the models in the downstream tasks at a better starting point. In this section, we first pre-train a GNN using the pre-training strategies designed in Section \ref{ssy1210:method}. Then, we fine-tune the model on different downstream tasks by using the parameters learned in the pre-training process. The downstream tasks we use here include the prediction of paper–field,  author name disambiguation (Author ND), and prediction of paper–venue. The OAG dataset contains published papers ranging from 1900 to 2019. In the following experiments, we divided the OAG dataset according to different time periods. Specifically, we use data before 2014 for pre-training and data after 2014 for fine-tuning. In the fine-tuning stage, we use the data from 2014 to 2016 as training data, the data in 2017 as validation data, and the data from 2018 to 2019 to test the performance of the models in downstream tasks. In order to be consistent with the fact that labeled data is usually scarce in downstream tasks, we only use 10$\%$ of the data in the fine-tuning step. The performance of all models is evaluated in terms of $\emph{Recall@10}$ $(R@10)$ and $\emph{NDCG@10}$ $(N@10)$.} 

\subsubsection{\textbf{Pre-training Baselines}} {In order to verify the superiority of the proposed PHE, we use the following methods capable of being extended to large-scale heterogeneous graphs as baselines:
\begin{itemize}[leftmargin=*]
    \item {\emph{No pre-train}. The model in the downstream task is initialized with random parameters instead of the parameters learned from the pre-training stage.}
    \item {\emph{GAE}~\cite{arxiv:ThomasMax16}. The core idea of the GAE is to mask a certain number of edges in the graph randomly and require the model to reconstruct these masked edges. Essentially, the GAE uses the traditional link prediction task to pre-train GNNs.}
    \item {\emph{un-SAGE}~\cite{DBLP:conf/nips/HamiltonYing17}. The un-SAGE is the GraphSAGE in an unsupervised setting, which encourages structurally adjacent nodes to have similar embeddings.}
    \item {\emph{DGI}~\cite{DBLP:conf/iclr/VelickovicFedus19}. The DGI aims to maximize the mutual information between global-level embeddings (graph summary embeddings) and local-level embeddings (local node embeddings).}
    \item {\emph{PTHGNN}~\cite{DBLP:conf/kdd/JiangJia21}. The PTHGNN is an approach that attempts to contrastively preserve semantic and structural properties in heterogeneous graphs by using contrastive learning.}
    \item {\emph{SimGRACE}~\cite{DBLP:conf/www/XiaWu22}. The SimGRACE is a pre-training method that relies on maximizing the mutual information between pairs of augmented data that share similar semantics.}
    \textcolor{black}{\item {\emph{GraphACL}~\cite{DBLP:conf/nips/XiaoZhu24}. GraphACL is an augmentation-free method, which aims to construct asymmetric views of neighboring nodes for contrastive learning.}}
\end{itemize}
\subsubsection{\label{ssy1210:implement_details} \textbf{Implementation Details}} {We adopt NVIDIA RTX A6000 (48GB GPU) and Intel(R) Xeon(R) Silver 4309Y CPU to conduct experiments on both the pre-training and downstream tasks. For a fair comparison, all baselines use the Transformer-based heterogeneous encoding as the base GNN. Considering that the heterogeneous graphs used in this paper are too large to be directly used for pre-training GNNs, we adopt a sampling strategy here to make the algorithm fit into the hardware. Specifically, we sample dense subgraphs for training~\cite{DBLP:conf/www/HuDong20}. We sample 128 nodes for each type per layer and repeat 6 times. For the base GNN, we set the number of layers as 3, the number of heads as 8, and the number of hidden units as 400. The AdamW optimizer with Cosine Annealing Learning Rate Scheduler with 100 epochs is used to pre-train the GNN. We choose the model that performs best on the validation data as the best pre-trained GNN. The lengths of the negative sample queues in Eq.~\ref{ssy1210:neg_sample_part2_query} and Eq.~\ref{ssy1210:neg_sample_positive_enhanced} are both 256. To keep the size of the queue equal to 256, we dynamically update the queue by adding the latest negative samples and removing the oldest negative samples (i.e., the negative samples in the latest $M$ batches). The temperature coefficients in Eq.~\ref{ssy1210:loss1} and Eq.~\ref{ssy1210:loss2} are set to $0.2$. The  hyper-parameter $\mu$ for the uniform distribution is tuned in $\{0.05,0.1,0.15,0.2\}$. The hyper-parameter $\lambda$ for balancing the effect of different pre-training tasks is tuned in $\{0.1,0.2,0.3,0.4,0.5\}$.}

\subsection{Performance Comparison (RQ1)}
In this section, we compare PHE to all baselines by conducting two types of experiments, i.e., the transfer experiment within the same domain and the transfer experiment across domains. The experimental results are summarized in Tables~\ref{ssy1210:withdomain} and \ref{ssy1210:crossdomain}.

\begin{table*}
\vspace{0.6cm}
\renewcommand\arraystretch{1.22}
  \centering
  \caption{Performance ($\%$) of different pre-training strategies in the transfer experiment within the same domain. The best results are marked in bold and the second-best results are underlined.}
  \scalebox{0.73}{
    \begin{tabular}{ccccccccccc}
    \toprule[1.5pt]
          & Task       & Metric       & no pre-train & un-SAGE & DGI & GAE & SimGRACE & PTHGNN & GraphACL & PHE \\
    \midrule
    \multirow{6}[1]{*}{MAT} 
            & \multirow{2}[1]{*}{Paper-Venue} 
                    & {N@10} & 31.15$_{\pm1.70}$ & 43.52$_{\pm0.27}$ & 38.47$_{\pm0.78}$ & 41.64$_{\pm1.25}$ & 42.73$_{\pm0.28}$ & \underline{44.27$_{\pm0.16}$} & 41.25$_{\pm0.09}$ & \textbf{45.30$_{\pm0.34}$} \\
            &       & {R@10} & 46.09$_{\pm1.69}$ & \underline{59.71$_{\pm0.53}$} & 54.54$_{\pm0.78}$ & 57.81$_{\pm0.10}$ & 59.02$_{\pm0.37}$ & 59.66$_{\pm0.60}$ & 56.02$_{\pm0.28}$ & \textbf{60.98$_{\pm0.60}$} \\
            & \multirow{2}[0]{*}{Paper-Field} 
                    & {N@10} & 19.40$_{\pm0.26}$ & 30.16$_{\pm0.27}$ & 22.99$_{\pm0.69}$ & 29.69$_{\pm0.32}$ & 29.35$_{\pm0.09}$ & \textbf{31.65$_{\pm0.06}$} & 29.96$_{\pm0.11}$ & {\underline{30.37$_{\pm0.05}$}} \\
            &       & {R@10} & 22.94$_{\pm0.24}$ & 33.78$_{\pm0.26}$ & 26.82$_{\pm0.69}$ & 33.12$_{\pm0.23}$ & 32.58$_{\pm0.04}$ & \textbf{35.00$_{\pm0.04}$} & 31.62$_{\pm0.31}$ & {\underline{33.79$_{\pm0.09}$}} \\
            & \multirow{2}[0]{*}{Author-ND} & {N@10} & 71.29$_{\pm1.11}$ & 71.84$_{\pm0.73}$ & 71.44$_{\pm0.69}$ & 72.35$_{\pm0.16}$ & \underline{72.57$_{\pm0.24}$} & 72.23$_{\pm0.64}$ & 72.30$_{\pm0.53}$ & \textbf{75.40$_{\pm1.32}$} \\
            &       & {R@10} & 93.33$_{\pm0.33}$ & 93.04$_{\pm0.20}$ & \underline{93.96$_{\pm1.58}$} & 93.04$_{\pm0.19}$ & 93.81$_{\pm0.53}$ & 93.32$_{\pm0.33}$ & 93.66$_{\pm0.14}$ & \textbf{94.23$_{\pm0.43}$} \\
    \midrule
    \multirow{6}[1]{*}{ENG} 
            & \multirow{2}[0]{*}{Paper-Venue} & {N@10} & 33.58$_{\pm1.88}$ & 41.30$_{\pm1.16}$ & 43.34$_{\pm0.13}$ & 43.61$_{\pm0.52}$ & 46.72$_{\pm0.45}$ & \underline{49.66$_{\pm0.73}$} & 40.83$_{\pm1.08}$ & \textbf{55.05$_{\pm0.28}$} \\
            &       & {R@10} & 48.55$_{\pm2.06}$ & 56.61$_{\pm1.31}$ & 59.96$_{\pm0.95}$ & 58.33$_{\pm1.01}$ & 62.47$_{\pm0.52}$ & \underline{65.42$_{\pm1.28}$} & 56.09$_{\pm1.31}$ & \textbf{69.66$_{\pm0.10}$} \\
            & \multirow{2}[0]{*}{Paper-Field} & {N@10} & 10.95$_{\pm0.22}$ & 21.60$_{\pm0.54}$ & 17.78$_{\pm0.52}$ & 20.04$_{\pm0.30}$ & \underline{23.33$_{\pm0.48}$} & 23.17$_{\pm0.10}$ & 23.07$_{\pm0.15}$ & \textbf{24.31$_{\pm0.43}$} \\
            &       & {R@10} & 13.32$_{\pm0.20}$ & 24.95$_{\pm0.56}$ & 20.76$_{\pm0.53}$ & 22.85$_{\pm0.35}$ & \underline{26.42$_{\pm0.27}$} & 26.30$_{\pm0.09}$ & 26.02$_{\pm0.20}$ & \textbf{27.26$_{\pm0.20}$} \\
            & \multirow{2}[1]{*}{Author-ND} & {N@10} & 76.96$_{\pm0.51}$ & \underline{79.17$_{\pm0.13}$} & 77.53$_{\pm0.48}$ & 77.86$_{\pm2.24}$ & 78.53$_{\pm0.37}$ & 78.93$_{\pm0.46}$ & 78.00$_{\pm0.81}$ & \textbf{80.81$_{\pm0.55}$} \\
            &       & {R@10} & 99.49$_{\pm0.01}$ & 99.44$_{\pm0.05}$ & 99.49$_{\pm0.01}$ & 99.49$_{\pm0.08}$ & 99.49$_{\pm0.01}$ & \underline{99.49$_{\pm0.01}$} & 99.21$_{\pm0.16}$ & \textbf{99.72$_{\pm0.13}$} \\
    \midrule
    \multirow{6}[2]{*}{CS} 
            & \multirow{2}[1]{*}{Paper-Venue} & {N@10} & 17.84$_{\pm0.04}$ & 28.65$_{\pm0.45}$ & 23.55$_{\pm1.20}$ & 28.41$_{\pm1.27}$ & 27.91$_{\pm0.37}$ & \textbf{34.02$_{\pm0.33}$} & 22.12$_{\pm1.01}$ & {\underline {31.28$_{\pm0.94}$}} \\
            &       & {R@10} & 28.58$_{\pm0.83}$ & 41.37$_{\pm0.61}$ & 34.86$_{\pm1.71}$ & 40.63$_{\pm1.70}$ & 40.79$_{\pm0.53}$ & \textbf{46.74$_{\pm0.13}$} & 32.94$_{\pm1.61}$ & {\underline{43.74$_{\pm1.57}$}} \\
            & \multirow{2}[0]{*}{Paper-Field} & {N@10} & 10.86$_{\pm0.18}$ & 16.57$_{\pm0.04}$ & 14.81$_{\pm0.24}$ & 19.16$_{\pm0.33}$ & 17.87$_{\pm0.33}$ & \underline{20.64$_{\pm0.15}$} & 18.73$_{\pm0.07}$ & \textbf{21.05$_{\pm0.06}$} \\
            &       & {R@10} & 12.60$_{\pm0.21}$ & 18.98$_{\pm0.06}$ & 17.22$_{\pm0.26}$ & 21.65$_{\pm0.31}$ & 20.13$_{\pm0.25}$ & \underline{23.07$_{\pm0.17}$} & 20.70$_{\pm0.18}$ & \textbf{23.24$_{\pm0.14}$} \\
            & \multirow{2}[1]{*}{Author-ND} & {N@10} & 77.18$_{\pm0.22}$ & 79.31$_{\pm0.23}$ & 77.95$_{\pm0.14}$ & \underline{80.08$_{\pm0.40}$} & 79.84$_{\pm1.19}$ & 78.96$_{\pm0.88}$ & 78.76$_{\pm0.49}$ & \textbf{81.21$_{\pm0.97}$} \\
            &       & {R@10} & 97.38$_{\pm0.07}$ & 96.71$_{\pm0.17}$ & 97.25$_{\pm0.02}$ & 97.41$_{\pm0.21}$ & {97.65$_{\pm0.40}$} & 96.96$_{\pm0.40}$ & \underline{97.72$_{\pm0.20}$} & \textbf{98.01$_{\pm0.05}$} \\
    \bottomrule[1.5pt]
    \end{tabular}}%
    \label{ssy1210:withdomain}
\end{table*}%

\begin{table*}[h]
\vspace{0.6cm}
\renewcommand\arraystretch{1.22}
  \centering
  \caption{Performance ($\%$) of different pre-training strategies in the transfer experiment across domains. The best results are marked in bold, and the second-best results are underlined.}
  \scalebox{0.74}{
    \begin{tabular}{ccccccccccc}
    \toprule[1.5pt]
          &Task       &Metric      & no pre-train & un-SAGE & DGI & GAE & SimGRACE & PTHGNN & GraphACL & PHE \\
    \midrule
    \multirow{6}[2]{*}{\makecell[c]{CS \\ $\downarrow$ \\ ENG}} & \multirow{2}[1]{*}{Paper-Venue} & {N@10} & 36.48$_{\pm0.55}$ & 38.77$_{\pm0.39}$ & 38.44$_{\pm1.90}$ & 38.15$_{\pm2.09}$ & 38.70$_{\pm0.62}$ & \underline{39.38$_{\pm2.05}$} & 32.88$_{\pm0.08}$ & \textbf{40.27$_{\pm0.33}$} \\
          &       & {R@10} & 51.95$_{\pm0.55}$ & 54.06$_{\pm0.85}$ & \underline{55.56$_{\pm1.95}$} & 53.07$_{\pm2.67}$ & 54.02$_{\pm1.19}$ & 53.70$_{\pm2.63}$ & 46.72$_{\pm1.60}$ & \textbf{56.08$_{\pm0.29}$} \\
          & \multirow{2}[0]{*}{Paper-Field} & {N@10} & 10.95$_{\pm0.22}$ & 16.37$_{\pm0.22}$ & 16.14$_{\pm0.34}$ & 17.35$_{\pm0.65}$ & 17.51$_{\pm0.67}$ & \underline{19.11$_{\pm0.32}$} & 17.18$_{\pm0.44}$ & \textbf{19.50$_{\pm0.05}$} \\
          &       & {R@10} & 13.32$_{\pm0.20}$ & 18.77$_{\pm0.33}$ & 18.91$_{\pm0.32}$ & 19.91$_{\pm0.75}$ & 20.19$_{\pm0.81}$ & \underline{22.23$_{\pm0.49}$} & 19.65$_{\pm0.74}$ & \textbf{22.51$_{\pm0.03}$} \\
          & \multirow{2}[1]{*}{Author-ND} & {N@10} & 76.96$_{\pm0.51}$ & 79.35$_{\pm0.35}$ & 79.00$_{\pm0.45}$ & 78.80$_{\pm0.33}$ & \underline{79.63$_{\pm0.20}$} & 79.42$_{\pm1.19}$ & 79.71$_{\pm0.60}$ & \textbf{80.32$_{\pm0.64}$} \\
          &       & {R@10} & 99.49$_{\pm0.01}$ & 99.55$_{\pm0.10}$ & 99.49$_{\pm0.01}$ & 99.55$_{\pm0.10}$ & \underline{99.55$_{\pm0.09}$} & 99.46$_{\pm0.05}$ & 99.33$_{\pm0.05}$ & \textbf{99.61$_{\pm0.13}$} \\
    \midrule
    \multirow{6}[2]{*}{\makecell[c]{ENG \\ $\downarrow$ \\ MAT}} & \multirow{2}[1]{*}{Paper-Venue} & {N@10} & 32.46$_{\pm1.80}$ & 38.68$_{\pm0.64}$ & 38.74$_{\pm0.60}$ & 35.07$_{\pm1.88}$ & 37.12$_{\pm0.17}$ & \underline{40.00$_{\pm0.27}$} & 26.67$_{\pm1.10}$ & \textbf{40.84$_{\pm0.43}$} \\
          &       & {R@10} & 47.08$_{\pm2.05}$ & 53.71$_{\pm1.09}$ & 53.63$_{\pm1.20}$ & 50.14$_{\pm1.98}$ & 53.21$_{\pm0.61}$ & \underline{55.27$_{\pm0.50}$} & 40.48$_{\pm1.38}$ & \textbf{55.63$_{\pm0.14}$} \\
          & \multirow{2}[0]{*}{Paper-Field} & {N@10} & 19.40$_{\pm0.26}$ & 23.63$_{\pm0.42}$ & 23.43$_{\pm0.08}$ & 23.63$_{\pm0.59}$ & 26.26$_{\pm0.35}$ & \textbf{27.92$_{\pm0.09}$} & 26.61$_{\pm0.23}$ & {\underline {26.91$_{\pm0.11}$}} \\
          &       & {R@10} & 22.94$_{\pm0.24}$ & 27.20$_{\pm0.52}$ & 27.25$_{\pm0.13}$ & 27.70$_{\pm0.66}$ & 29.69$_{\pm0.34}$ & \textbf{31.62$_{\pm0.17}$} & 30.35$_{\pm0.15}$ & {\underline {30.72$_{\pm0.04}$}} \\
          & \multirow{2}[1]{*}{Author-ND} & {N@10} & 71.29$_{\pm1.11}$ & 70.77$_{\pm0.51}$ & \underline{72.25$_{\pm0.35}$} & 71.13$_{\pm0.53}$ & 71.51$_{\pm0.77}$ & 70.45$_{\pm0.39}$ & 71.08$_{\pm0.42}$ & \textbf{75.39$_{\pm0.08}$} \\
          &       & {R@10} & 93.33$_{\pm0.33}$ & \underline{93.59$_{\pm0.23}$} & 93.35$_{\pm0.23}$ & 92.67$_{\pm0.05}$ & 93.38$_{\pm0.28}$ & 93.22$_{\pm0.19}$ & 92.91$_{\pm0.24}$ & \textbf{94.42$_{\pm0.27}$} \\
    \bottomrule[1.5pt]
    \end{tabular}}%
    \label{ssy1210:crossdomain}
\end{table*}%

In the transfer experiment within the same domain (as shown in Table~\ref{ssy1210:withdomain}), the proposed PHE demonstrates certain superiority over its counterparts. First, the PHE achieves an average performance gain of 46.5$\%$ compared to the base GNN model without pre-training, which illustrates that the PHE can help learn transferable knowledge that is beneficial to downstream tasks. Essentially, the learned transfer knowledge enables the parameters of the model to be better initialized, thereby improving the performance of the model in the downstream task. Second, PHE outperforms other pre-training baselines and always ranks the top two among different downstream tasks on different datasets. The reason is that PHE can capture high-order information in the heterogeneous graph (using the ${\mathbf S}_{1}$ to help capture heterogeneous information) and guide the model to learn knowledge with better transferability (using the ${\mathbf S}_{2}$ to help handle the problem of semantic mismatch). It is easy to see that the PTHGNN performs the second best. The good performance of the PTHGNN benefits from its utilization of different types of relations in the heterogeneous graph during the design of the node-level pre-training task. However, the PTHGNN assumes that different types of nodes have the same contribution during the design of the high-level pre-training task, which weakens its ability to acquire fine-grained heterogeneous information.

We also conduct the transfer experiment across domains to further verify whether the proposed method has the ability to learn transferable knowledge. The results in Table \ref{ssy1210:crossdomain} demonstrate that the proposed method has better performance than the other alternatives, which is consistent with the results in Table \ref{ssy1210:withdomain}. Specifically, the proposed method always ranks the top two among different downstream tasks on different datasets while its counterparts only perform well in some specific situations. All the above observations indicate that the proposed pre-training strategies can help capture high-order heterogeneous information in the graph and guide the model to learn knowledge with better transferability, thus leading to the performance gain of the PHE over the other baselines.

\subsection{Visualization and Analysis (RQ2)}
In this section, we first visualize the uniformity of the learned embeddings. Secondly, we show the flexible
architecture generalization of the proposed method. Thirdly, we provide a performance analysis of our proposed method under different fine-tuning modes. Furthermore, we compare the training time of the proposed method with that of its competitors. Finally, we present a sensitivity analysis of the model parameters.
\subsubsection{\textbf{Visualizing Embedding Distributions}}
\begin{figure}[htbp]
    \centering
    \subfloat[un-SAGE]{
        \includegraphics[width=1.25in]{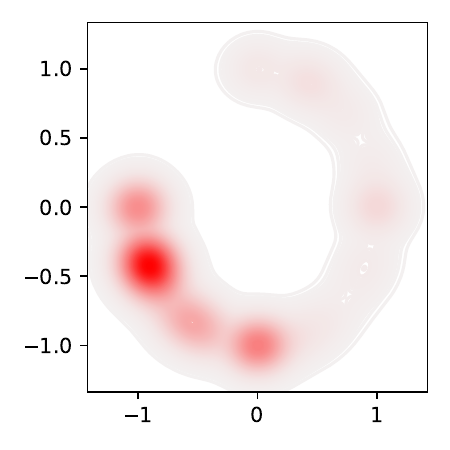}
        \label{label_for_cross_ref_1}
    }
    \subfloat[DGI]{
	\includegraphics[width=1.25in]{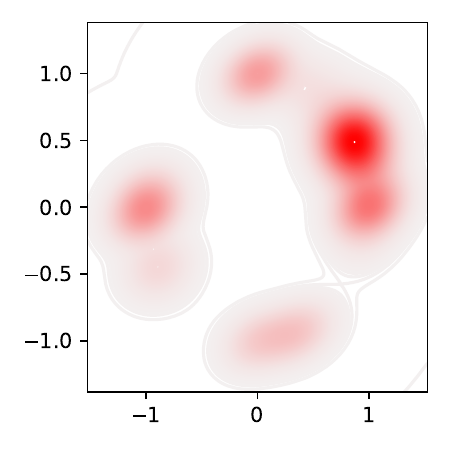}
        \label{label_for_cross_ref_2}
    }
    \subfloat[GAE]{
	\includegraphics[width=1.25in]{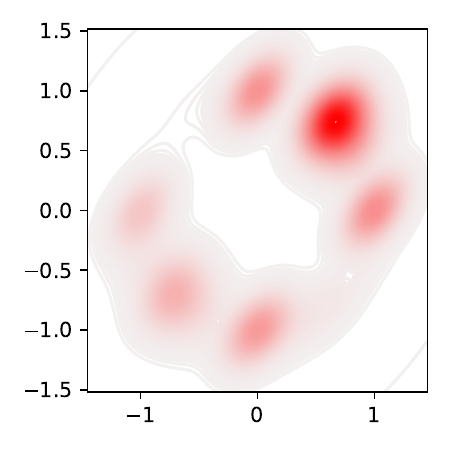}
        \label{label_for_cross_ref_2}
    }
    \vspace{15pt} 
    \subfloat[SimGRACE]{
        \includegraphics[width=1.25in]{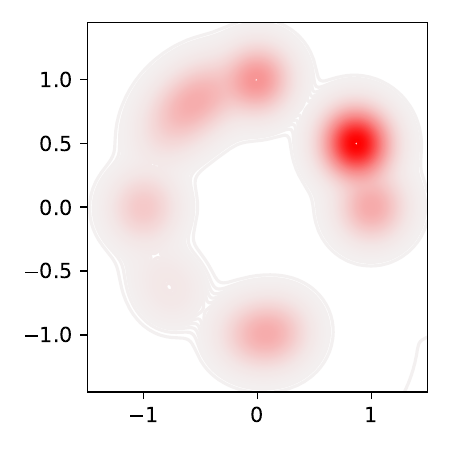}
        \label{label_for_cross_ref_3}
    }
    \subfloat[PTHGNN]{
	\includegraphics[width=1.25in]{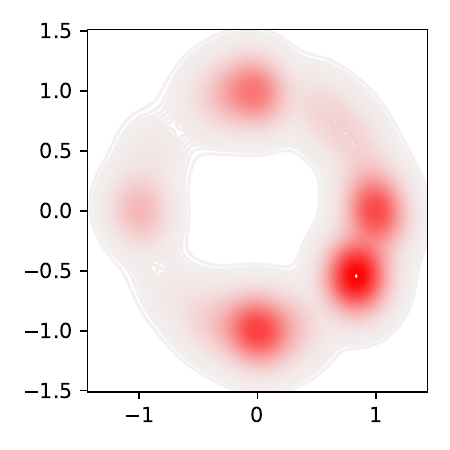}
        \label{label_for_cross_ref_4}
    }
    \subfloat[GraphACL]{
	\includegraphics[width=1.25in]{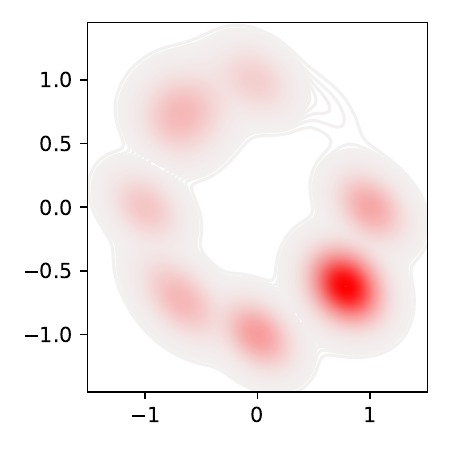}
        \label{label_for_cross_ref_5}
    }
    \subfloat[PHE]{
	\includegraphics[width=1.25in]{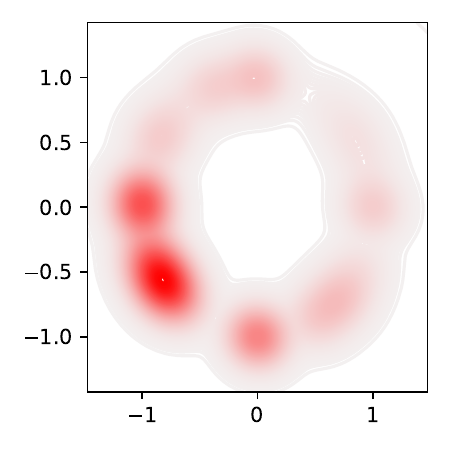}
        \label{label_for_cross_ref_6}
    }
    \caption{Visualizing the distribution of embeddings.}
    \label{ssy1210:KDE}
\end{figure}
In order to better understand the benefits of the proposed PHE, we plot embedding distributions with Gaussian kernel density estimation in this section. It has been shown that contrastive learning is closely related to the uniformity of learned embeddings~\cite{DBLP:conf/icml/WangIsola20,DBLP:journals/CMPB/WangWang25,DBLP:journals/CVM/ZhangChen24}, where the feature distribution that retains the maximal information about the embeddings is preferred. As shown in Fig. \ref{ssy1210:KDE}, the embedding learned by the proposed PHE demonstrates a more uniform distribution compared to its counterparts, which means that the proposed PHE can capture diverse information in the heterogeneous graph. 

\begin{table*}
\vspace{0.6cm}
\renewcommand\arraystretch{1.22}
  \centering
  \caption{\textcolor{black}{Performance ($\%$) of different pre-training strategies in the transfer experiments within the same domain and across domains (with the GCN architecture). The best results are marked in bold, and the second-best results are underlined.}}
  \scalebox{0.74}{
    \begin{tabular}{ccccccccccc}
    \toprule[1.5pt]
          &Task       &Metric      & no pre-train & un-SAGE & DGI & GAE & SimGRACE & PTHGNN & GraphACL & PHE \\
    \midrule
    \multirow{6}[2]{*}{\makecell[c]{MAT \\ $\downarrow$ \\ MAT}} & \multirow{2}[1]{*}{Paper-Venue} & {N@10} & 31.86$_{\pm0.20}$ & 33.29$_{\pm0.59}$ & 33.17$_{\pm0.72}$ & \underline{34.56$_{\pm0.05}$} & 33.68$_{\pm0.09}$ & 34.11$_{\pm1.67}$ & 34.71$_{\pm0.05}$ & \textbf{36.98$_{\pm0.12}$} \\
          &       & {R@10} & 46.48$_{\pm0.33}$ & 47.68$_{\pm0.93}$ & 47.85$_{\pm1.44}$ & \underline{50.34$_{\pm0.20}$} & 48.03$_{\pm0.11}$ & 48.96$_{\pm2.23}$ & 50.05$_{\pm0.27}$ & \textbf{51.80$_{\pm0.14}$} \\
          & \multirow{2}[0]{*}{Paper-Field} & {N@10} & 22.68$_{\pm0.07}$ & 24.86$_{\pm0.03}$ & 24.75$_{\pm0.18}$ & 25.32$_{\pm0.07}$ & 24.87$_{\pm0.04}$ & \underline{27.20$_{\pm0.11}$} & 28.11$_{\pm0.05}$ & \textbf{28.58$_{\pm0.18}$} \\
          &       & {R@10} & 26.29$_{\pm0.11}$ & 28.52$_{\pm0.09}$ & 28.48$_{\pm0.21}$ & 28.95$_{\pm0.10}$ & 28.79$_{\pm0.10}$ & \underline{30.80$_{\pm0.19}$} & 31.97$_{\pm0.10}$ & \textbf{32.52$_{\pm0.19}$} \\
          & \multirow{2}[1]{*}{Author-ND} & {N@10} & 69.78$_{\pm0.78}$ & 70.99$_{\pm0.81}$ & 70.62$_{\pm0.48}$ & 71.55$_{\pm0.12}$ & 70.16$_{\pm0.14}$ & \underline{71.63$_{\pm0.14}$} & 70.19$_{\pm0.22}$ & \textbf{71.86$_{\pm0.13}$} \\
          &       & {R@10} & 93.96$_{\pm0.19}$ & 94.59$_{\pm0.20}$ & \underline{94.60$_{\pm0.11}$} & 94.60$_{\pm0.20}$ & 93.83$_{\pm0.06}$ & 94.41$_{\pm0.11}$ & 92.85$_{\pm0.11}$ & \textbf{94.89$_{\pm0.06}$} \\
    \midrule
    \multirow{6}[2]{*}{\makecell[c]{MAT \\ $\downarrow$ \\ ENG}} & \multirow{2}[1]{*}{Paper-Venue} & {N@10} & 33.32$_{\pm0.07}$&35.94$_{\pm0.18}$&34.98$_{\pm0.27}$&35.95$_{\pm1.14}$&34.35$_{\pm0.53}$&3\underline{6.19$_{\pm0.88}$} & 37.52$_{\pm0.47}$ &\textbf{40.66$_{\pm0.15}$} \\
          &       & {R@10} & 50.23$_{\pm1.84}$&51.38$_{\pm0.16}$&50.66$_{\pm0.27}$&\underline{53.36$_{\pm0.59}$}&50.48$_{\pm0.51}$&52.68$_{\pm1.73}$ & 55.03$_{\pm0.80}$ &\textbf{57.96$_{\pm0.24}$} \\
          & \multirow{2}[0]{*}{Paper-Field} & {N@10} & 15.55$_{\pm0.30}$&16.68$_{\pm0.04}$&15.94$_{\pm0.13}$&15.83$_{\pm0.19}$&17.42$_{\pm0.11}$&\underline{18.55$_{\pm0.12}$}& 18.76$_{\pm0.02}$ &\textbf{19.66$_{\pm0.10}$} \\
          &       & {R@10} & 18.38$_{\pm0.32}$&19.73$_{\pm0.10}$&18.92$_{\pm0.11}$&18.62$_{\pm0.17}$&20.34$_{\pm0.15}$&\underline{21.57$_{\pm0.13}$}& 21.73$_{\pm0.08}$ &\textbf{22.56$_{\pm0.10}$} \\
          & \multirow{2}[1]{*}{Author-ND} & {N@10} & 78.84$_{\pm0.70}$&78.91$_{\pm0.41}$&79.69$_{\pm0.67}$&78.31$_{\pm0.85}$&\underline{80.79$_{\pm0.01}$}&75.18$_{\pm0.60}$& 76.98$_{\pm1.25}$ &\textbf{81.14$_{\pm0.02}$} \\
          &       & {R@10} & 99.23$_{\pm0.30}$&99.23$_{\pm0.01}$&\underline{99.40$_{\pm0.44}$}&99.23$_{\pm0.23}$&99.23$_{\pm0.01}$&99.00$_{\pm0.05}$& 99.63$_{\pm0.10}$ &\textbf{99.57$_{\pm0.01}$} \\
    \bottomrule[1.5pt]
    \end{tabular}}%
    \label{ssy1210:flexible_arch_genral}
\end{table*}%

\textcolor{black}{\subsubsection{\textbf{Flexible Generalization}}
In this experiment, we demonstrate the flexible generalization of the proposed method. The experiment on the the flexible architecture generalization is shown in Table \ref{ssy1210:flexible_arch_genral}, we utilized the well-known GCN~\cite{DBLP:conf/iclr/Thomas16} as the node encoding architecture instead of the transformer-based heterogeneous encoding in the transfer experiments within the same domain and across domains. The results in Table \ref{ssy1210:flexible_arch_genral} indicate that the proposed method consistently outperforms the comparison methods across multiple settings. Compared to the second-best method, the relative performance gain of the proposed method can reach up to $12.35\%$. We also test the generalization of the proposed method on other types of graph data (i.e., Reddit \cite{DBLP:conf/nips/HamiltonYL17}). Experimental results in Fig. \ref{ssy1210:flexible_data_genral} show that the proposed method outperforms other baselines. This demonstrates that PHE is not limited to academic networks and can be easily extended to other types of networks, such as social networks.}

\begin{figure*}
    \centering
    \begin{subfigure}[b]{0.9\linewidth}
        \includegraphics[width=\textwidth]{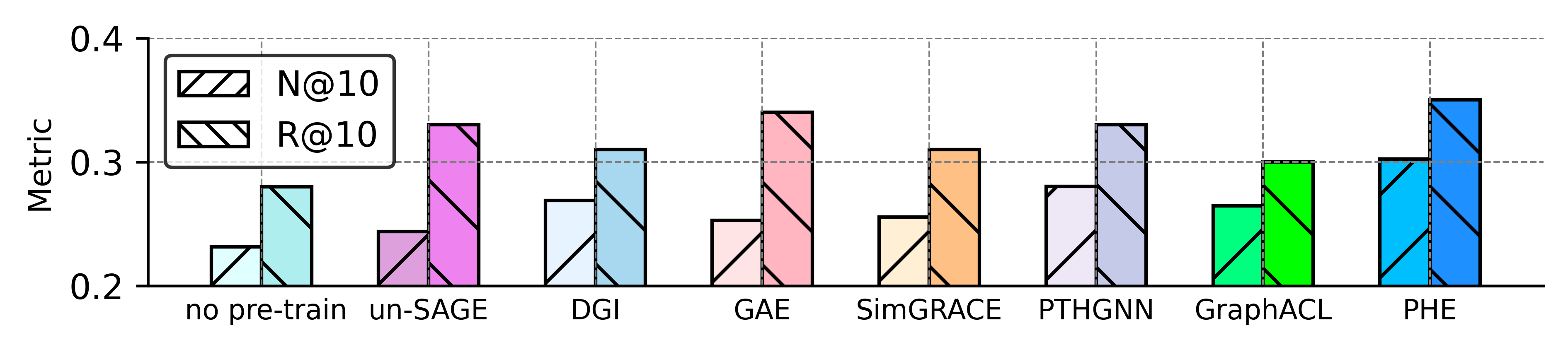}
        \caption{Fine-tuning with 50\% of labeled data.}
        \label{fig:a}
    \end{subfigure}
    \begin{subfigure}[b]{0.9\linewidth}
        \includegraphics[width=\textwidth]{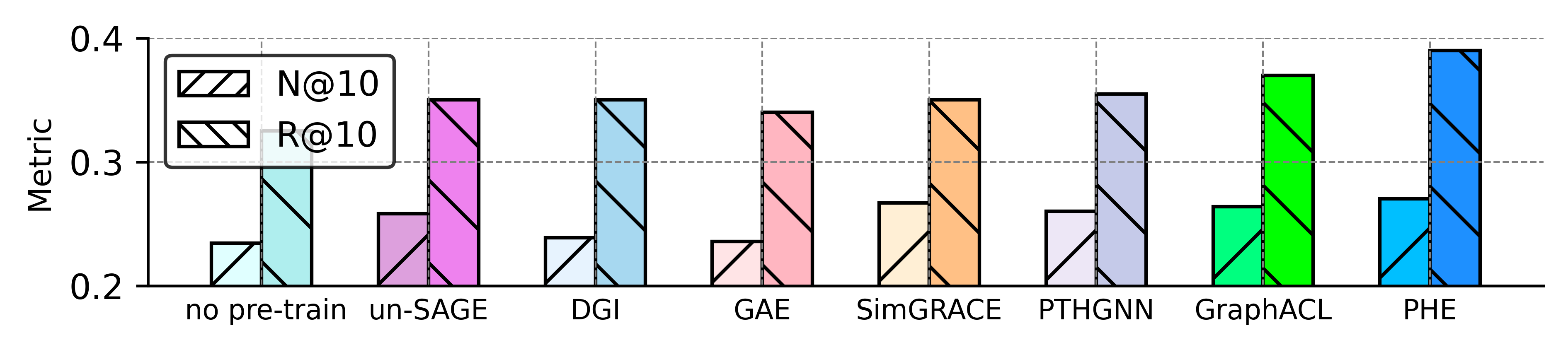}
        \caption{Fine-tuning with 100\% of labeled data.}
        \label{fig:b}
    \end{subfigure}
    \caption{Experiments on the generalization of the proposed method on other types of graph data (i.e., Reddit)}
    \label{ssy1210:flexible_data_genral}
\end{figure*}

\textcolor{black}{\subsubsection{\textbf{Training Time Analysis}}
In Section \ref{ssy1210:complexity}, we provide a detailed analysis of the model complexity. Compared with the best-performing baseline method PTHGNN (whose model complexity is offered in Appendix of \cite{DBLP:conf/kdd/JiangJia21}), we conclude that the complexity of the proposed PHE is on par with that of PTHGNN, residing within the same order of magnitude. The training time cost per epoch of different methods on various datasets is presented in Fig. \ref{ssy1210:training_time}. As shown in Fig. \ref{ssy1210:training_time}, the proposed method exhibits a comparable training time cost to that of the baseline methods.
}

\begin{figure*}
    \includegraphics[width=0.38\textwidth]{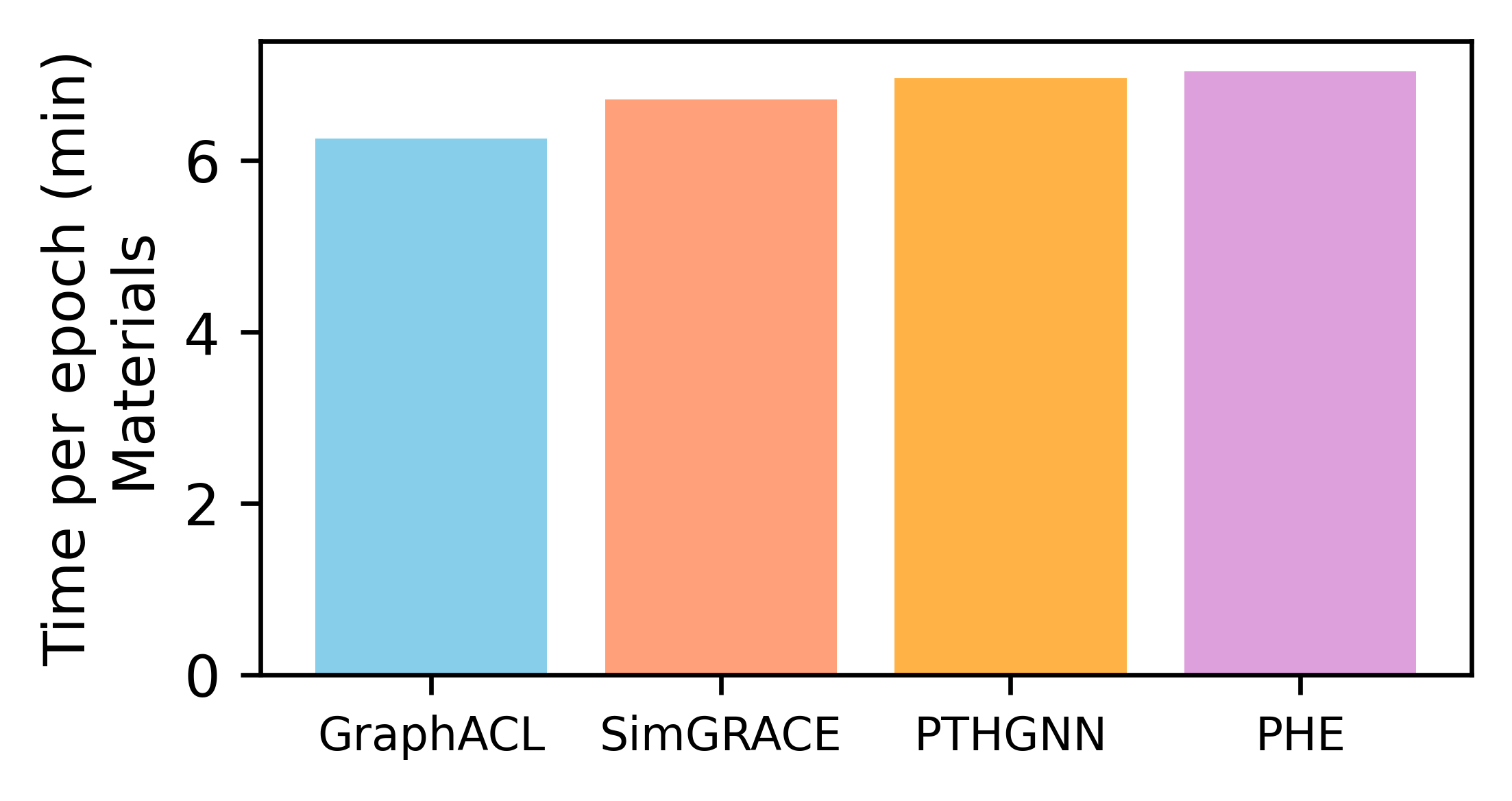}\hspace{10pt}
    \includegraphics[width=0.38\textwidth]{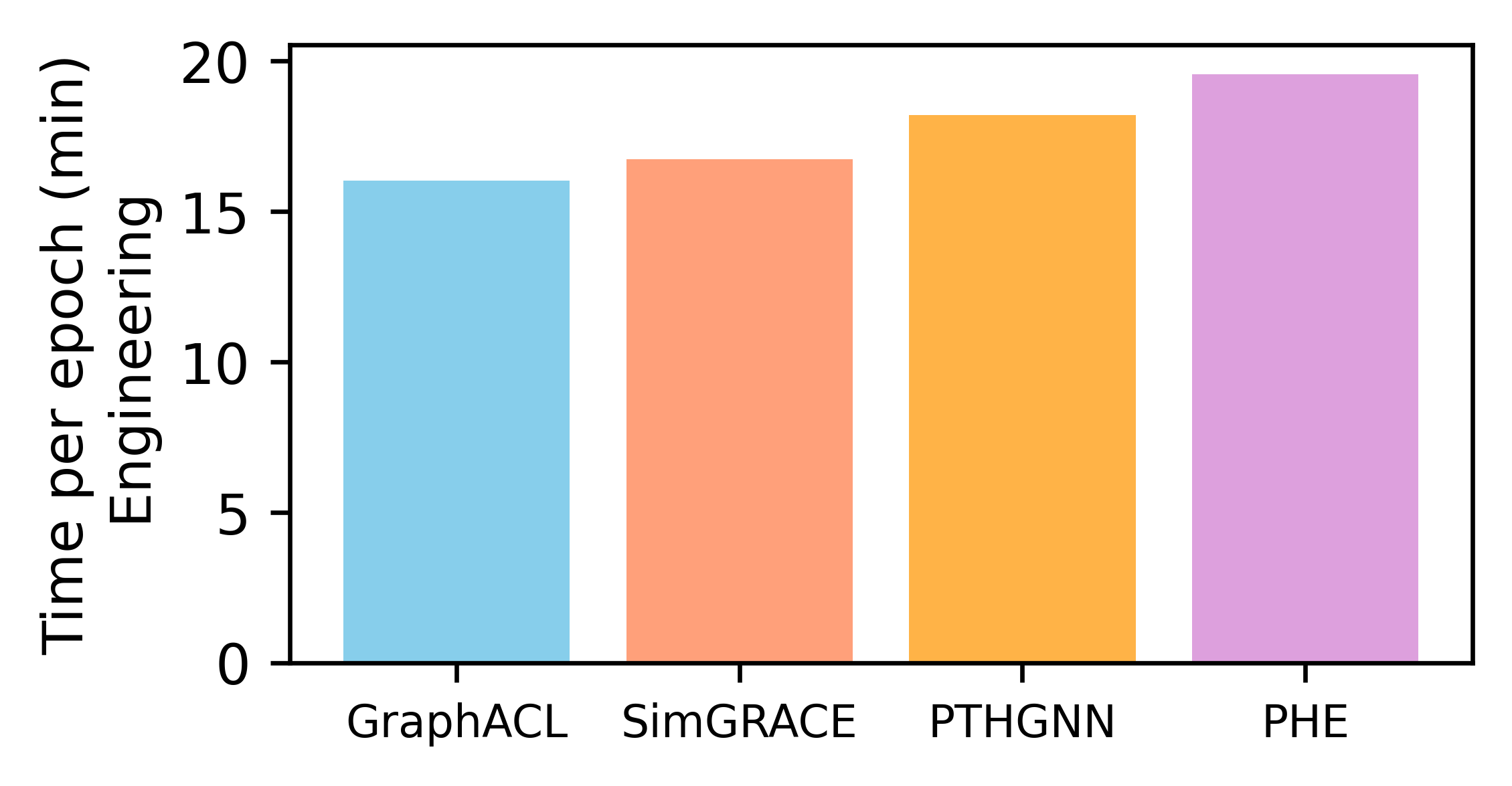}\\
    \includegraphics[width=0.38\textwidth]{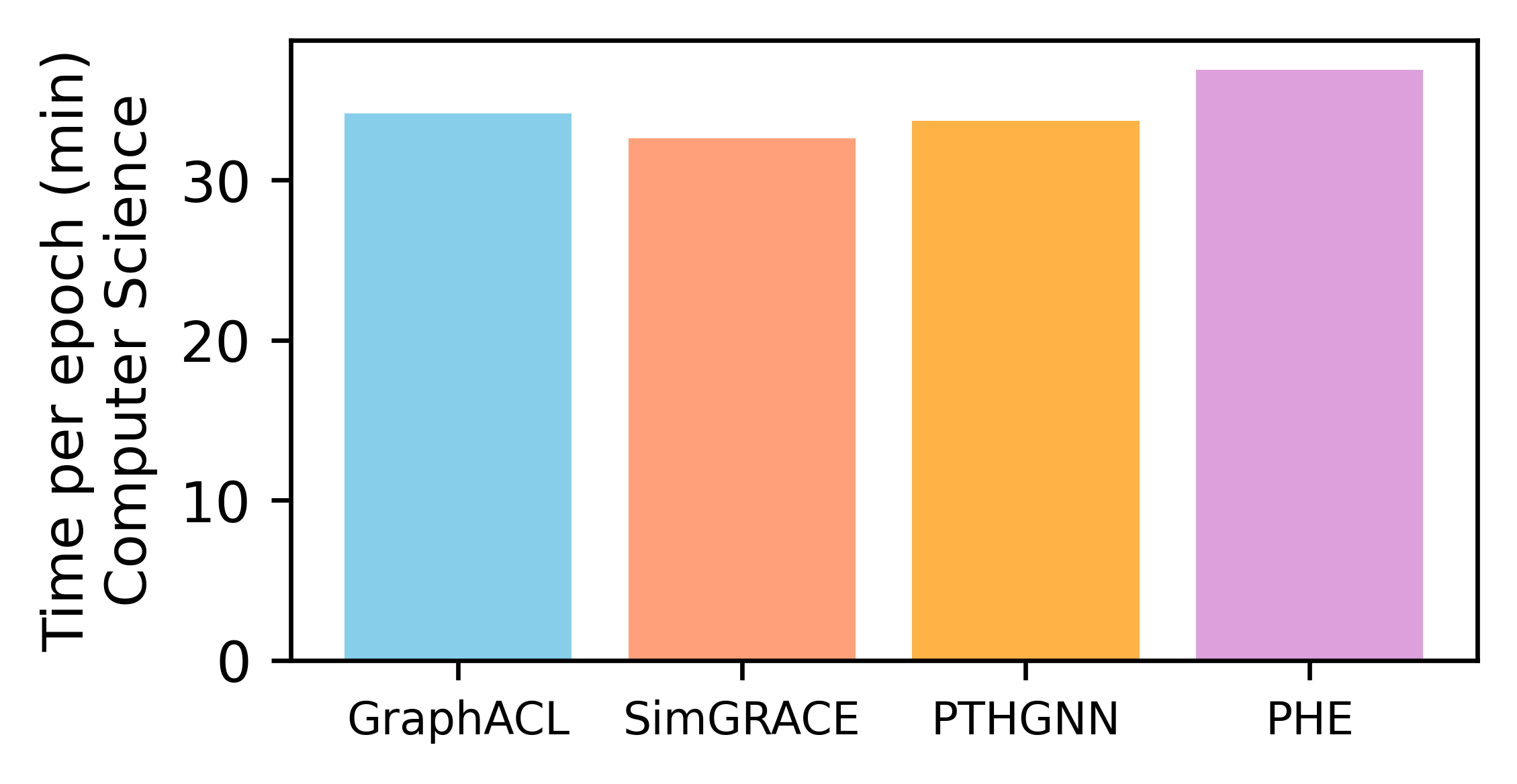}\hspace{10pt}
    \includegraphics[width=0.38\textwidth]{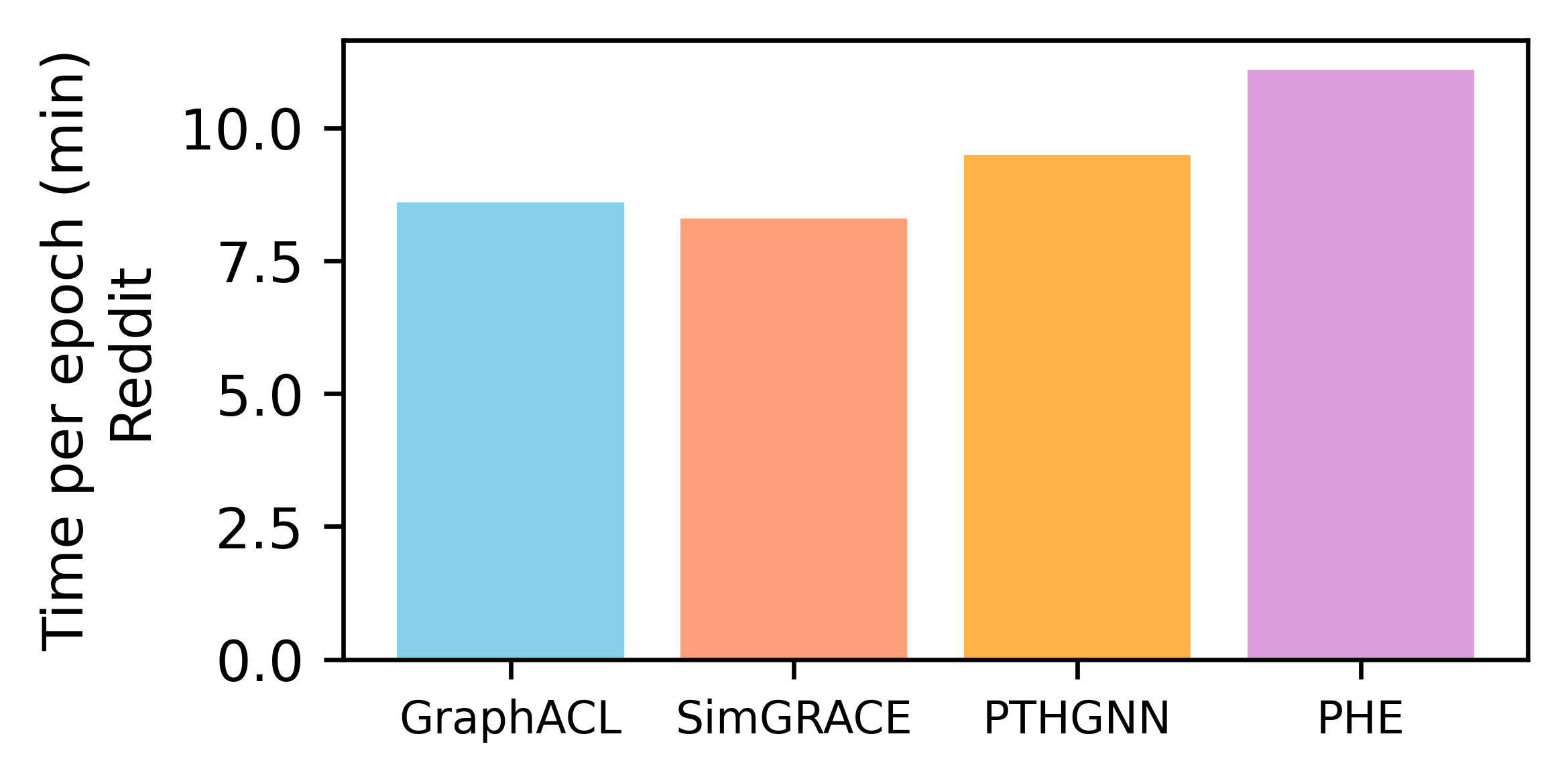}
    \caption{Training time cost per epoch of different methods on different datasets.}
    \label{ssy1210:training_time}
\end{figure*}

\subsubsection{\textbf{Frozen and Full Fine-tuning}}
In this experiment, two fine-tuning modes are used to evaluate whether the proposed method can learn transferable knowledge, namely full fine-tuning mode (FU-PHE) and frozen fine-tuning mode (FE-PHE). In the full fine-tuning mode, the pre-trained GNN and the classifier in the downstream task are trained together. In frozen fine-tuning mode, we freeze the parameters of the pre-trained GNN and only train the classifier in the downstream task. 

The performance comparison in three different modes is shown in Table \ref{ssy1210:freezetuning}. It can be seen that FE-PHE performs better than the model in the no pre-train mode, and the performance of FE-PHE is comparable to that of FU-PHE. All the above observations suggest that the proposed method can guide the model to capture transferable knowledge during the pre-training stage.

\begin{table}[htbp]
\renewcommand\arraystretch{1.0}
  \centering
  \caption{Performance of the PHE in different modes.}
  \scalebox{1.0}{
    \begin{tabular}{cccccccc}
    \toprule[1.5pt]
    \multicolumn{2}{c}{Task} & \multicolumn{2}{c}{Paper-Venue} & \multicolumn{2}{c}{Paper-Field} & \multicolumn{2}{c}{Author ND} \\
    \midrule
    \multicolumn{2}{c}{Metric} & {N@10} & {R@10} & {N@10} & {R@10} & {N@10} & {R@10} \\
    \midrule
    \multicolumn{2}{c}{no-pretrain} & 31.15 & 46.09 & 19.40  & 22.94 & 71.29 & 93.33 \\
    \multicolumn{2}{c}{FE-PHE} & 39.11 & 55.39 & 28.03 & 31.98 & 72.98 & 93.56 \\
    \multicolumn{2}{c}{FU-PHE} & 45.30 & 60.98 & 30.37  & 33.79 & 75.40 & 94.23 \\
    \bottomrule[1.5pt]
    \end{tabular}%
    }
  \label{ssy1210:freezetuning}%
\end{table}%

\subsubsection{\textbf{Parameter Sensitivity}}
In this section, we investigate the parameter sensitivity of the GNN architecture, the balancing coefficient $\lambda$, and the embedding size. The results are shown in Fig. \ref{ssy1210:backbone_para_diff} to Fig. \ref{ssy1210:dim_para_diff}. In this experiment, we first apply the proposed pre-training strategies to four different GNN architectures, i.e., GCN, SAGE, HAT, and Transformer-based heterogeneous encoding used in this paper (HGT). As shown in Fig. \ref{ssy1210:backbone_para_diff}, the pre-trained model with HGT layers achieves the best performance. This is because the HGT is specially designed for heterogeneous graphs, which can better adapt to the unique structure of heterogeneous graphs (multiple types of edges/nodes) and capture useful semantic information. Then, we investigate the parameter sensitivity of the balancing coefficient $\lambda$ in  Eq.~\ref{ssy1210:loss}. As shown in Fig. \ref{ssy1210:dim_para_lambda}, the performance is relatively stable across different choices of $\lambda$ and the proposed method achieves better performance when the value of $\lambda$ is around 0.4. Finally, we test the parameter sensitivity of
the embedding size. As shown in Fig.~\ref{ssy1210:dim_para_diff}, the performance is consistent across different choices of embedding size. \textcolor{black}{The parameter sensitivity of perturbation amplitude and temperature coefficient are provided in Fig. \ref{ssy1210:a_t}. It can be seen that both larger and smaller perturbation amplitudes lead to a decrease in performance. Additionally, the optimal value of the temperature parameter for the proposed model is 0.2.}
\begin{figure}[h]
    \centering
    \includegraphics[width=3.2in]{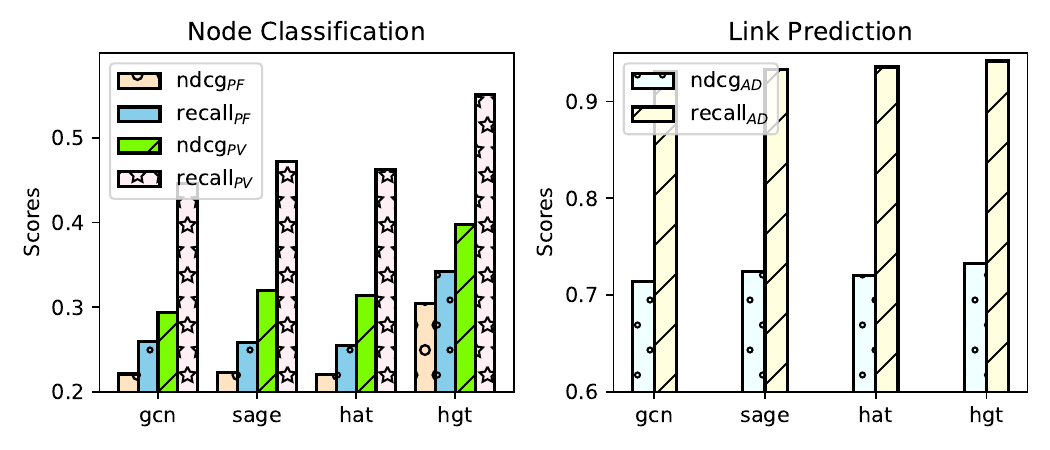}
    \caption{Parameter sensitivity of the GNN architecture.}
    \label{ssy1210:backbone_para_diff}
\end{figure}

\begin{figure}[h]
    \centering
    \includegraphics[width=3.2in]{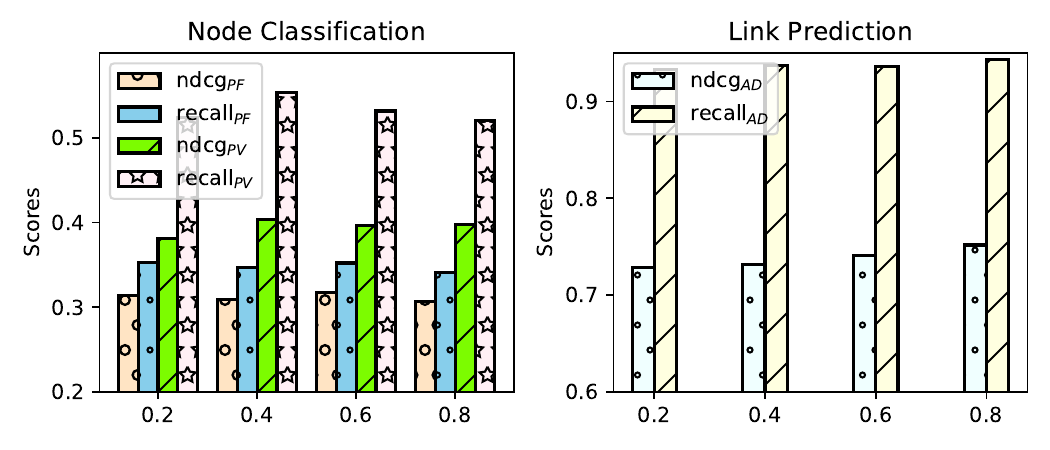}
    \caption{Parameter sensitivity of $\lambda$.}
    \label{ssy1210:dim_para_lambda}
\end{figure}

\begin{figure}[h]
    \centering
    \includegraphics[width=3.2in]{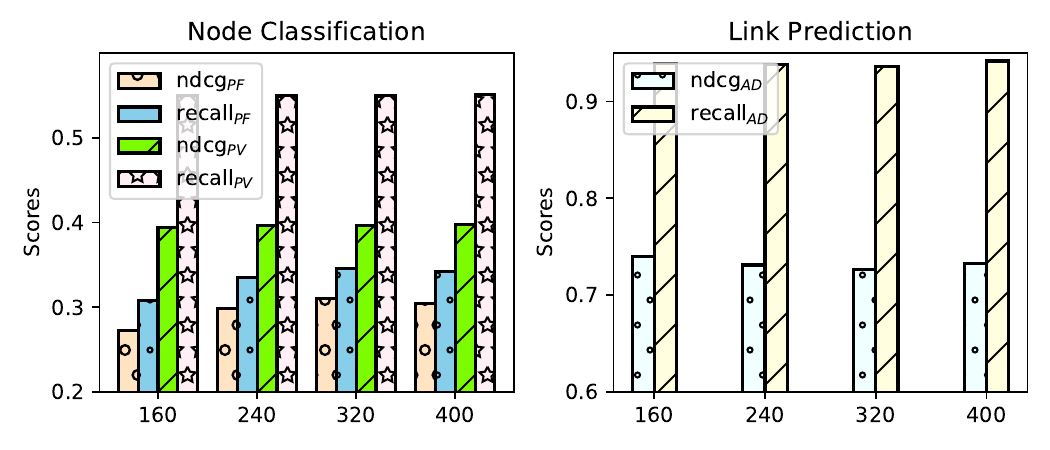}
    \caption{Parameter sensitivity of the embedding size.}
    \label{ssy1210:dim_para_diff}
\end{figure}

\begin{figure}[h]
    \centering
    \includegraphics[width=1.38in]{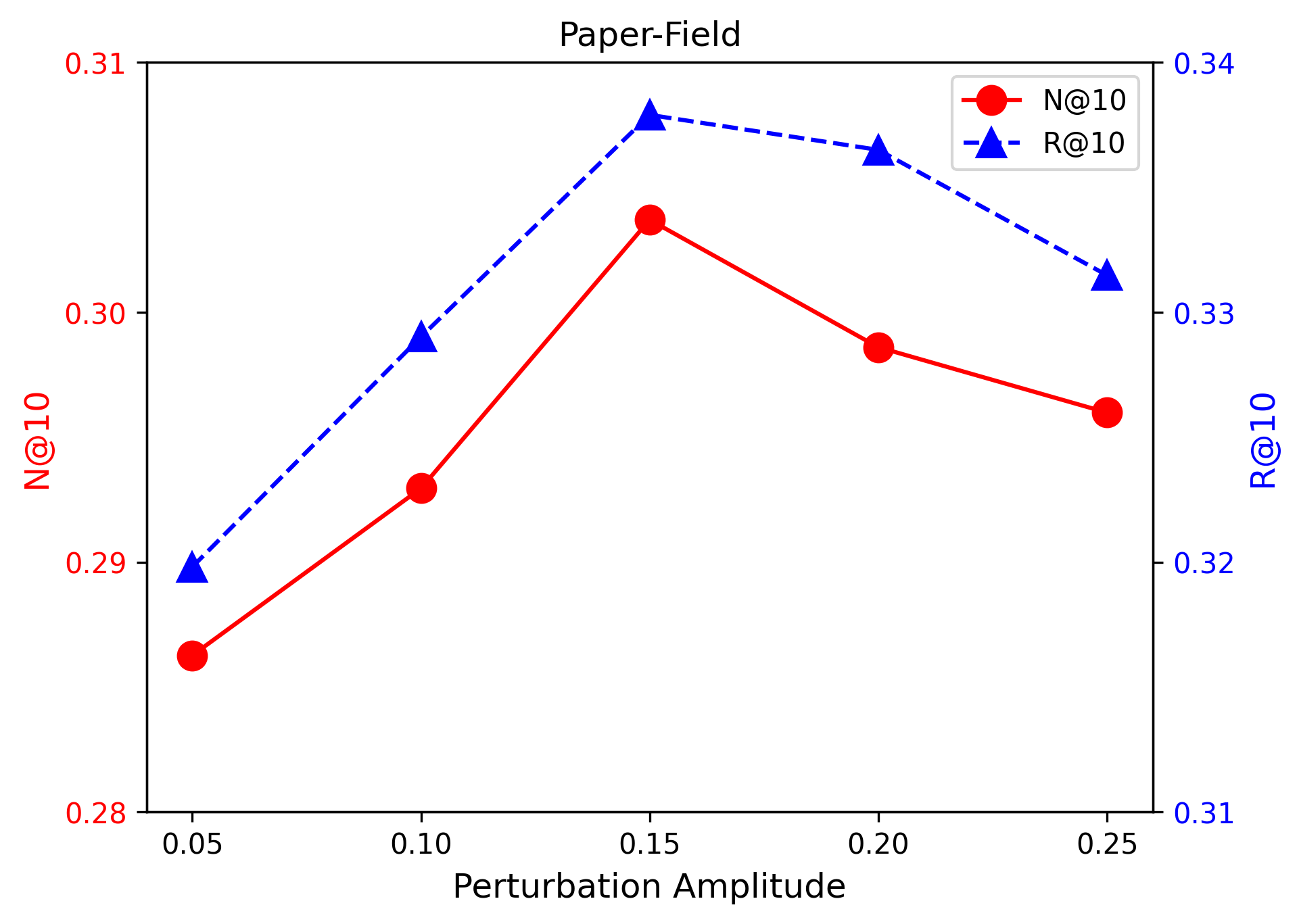}
    \includegraphics[width=1.38in]{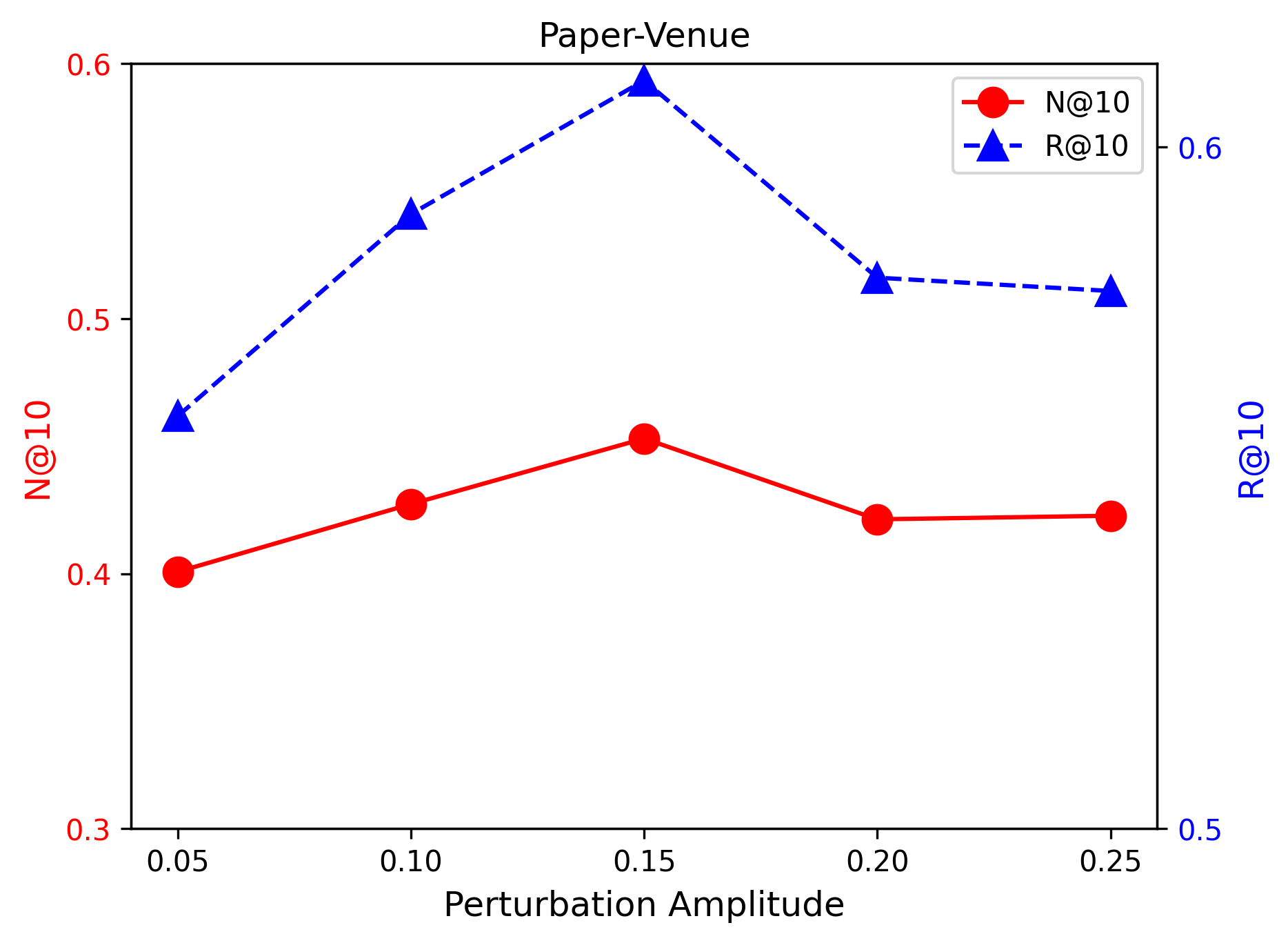}\\
    \includegraphics[width=1.38in]{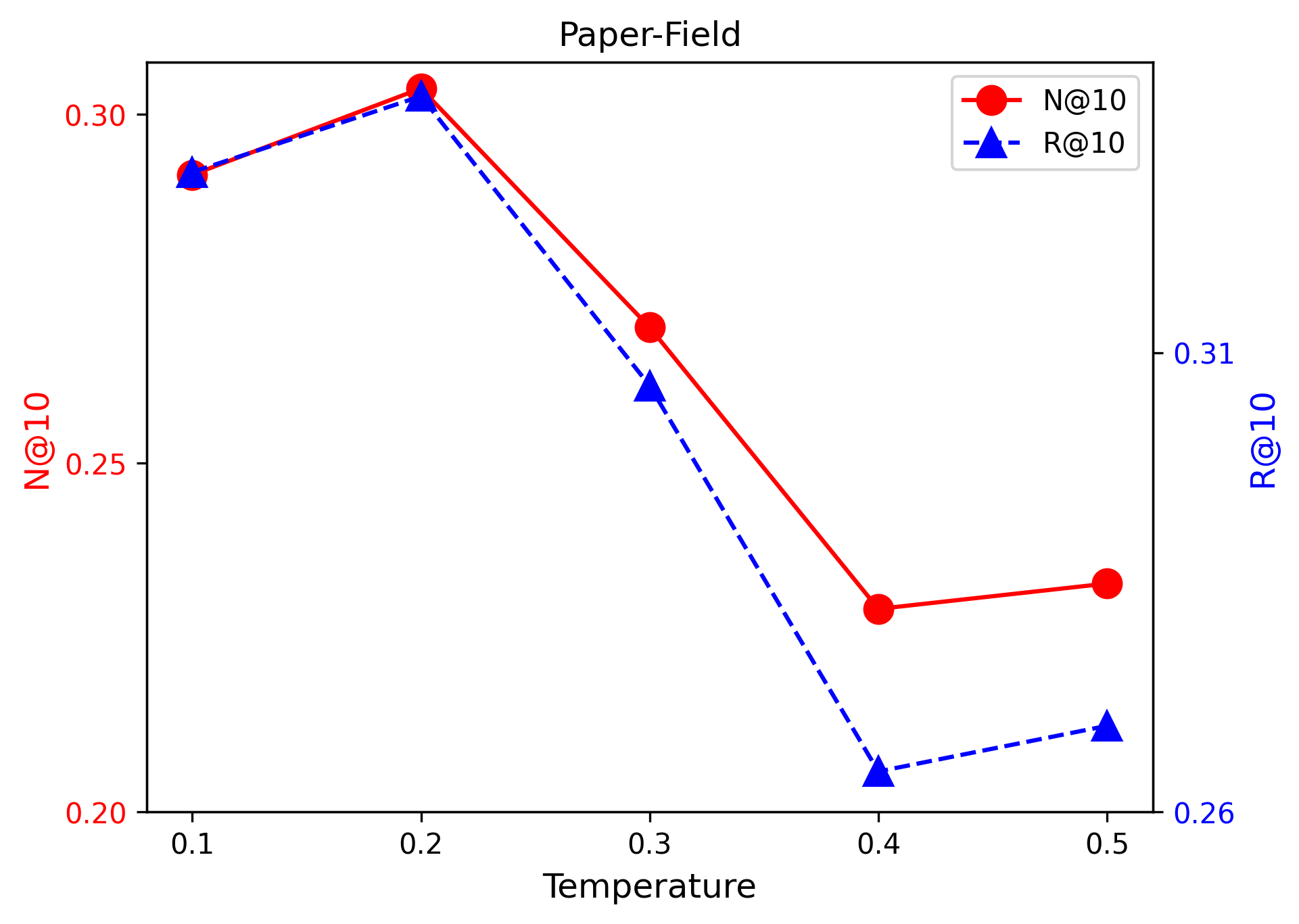}
    \includegraphics[width=1.38in]{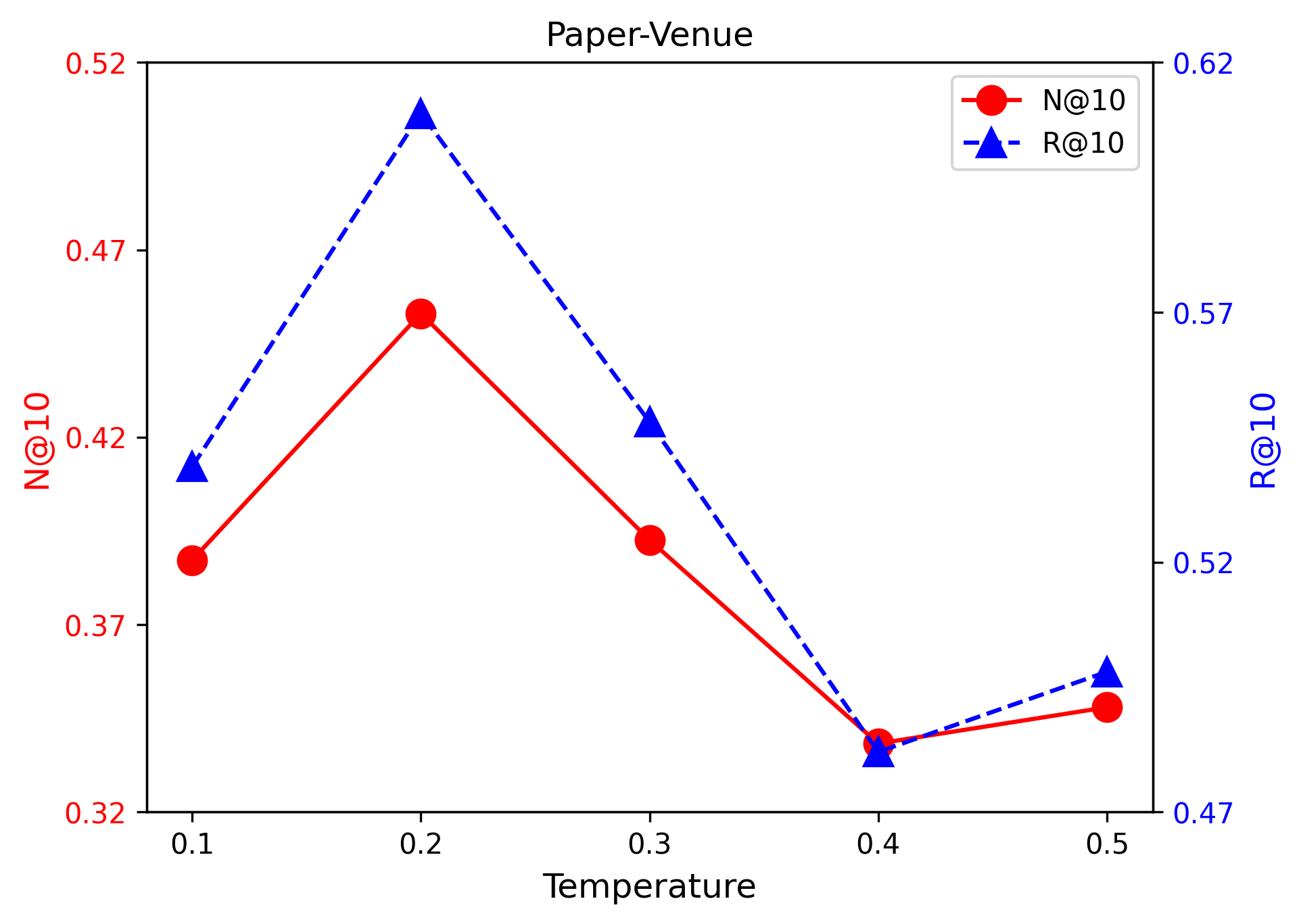}
    \caption{Parameter sensitivity of perturbation amplitude and temperature coefficient.}
    \label{ssy1210:a_t}
\end{figure}

\begin{table}[htbp]
\renewcommand\arraystretch{1.0}
  \centering
  \caption{Impact of different components ($pre, en1, en2$) in the transfer experiment within the same domain.}
  \scalebox{0.9}{
  \begin{tabular}{ccccccccc}
    \toprule[1.5pt]
    \multicolumn{3}{c}{CS$\rightarrow$CS} & \multicolumn{2}{c}{Paper-Venue} & \multicolumn{2}{c}{Paper-Field} & \multicolumn{2}{c}{Author ND} \\
    \midrule
    pre & en1 & en2 & {N@10} & {R@10} & {N@10} & {R@10} & {N@10} & {R@10} \\
    \midrule
    \textbf{\ding{56}} & \textbf{\ding{56}} & \textbf{\ding{56}} & 17.84 & 28.58 & 10.86 & 12.60 & 77.18 & 97.38 \\
    \textbf{\ding{52}} & \textbf{\ding{56}} & \textbf{\ding{56}} & 27.43 & 38.52 & 18.72 & 21.33 & 80.01 & 96.72 \\
    \textbf{\ding{52}} & \textbf{\ding{52}} & \textbf{\ding{56}} & 30.06 & 38.09 & 20.20  & 22.56 & 80.68 & 97.72 \\
    \textbf{\ding{52}} & \textbf{\ding{56}} & \textbf{\ding{52}} & 28.70 & 40.12 & 19.20 & 21.40 & 80.49 & 97.81 \\
    \textbf{\ding{52}} & \textbf{\ding{52}} & \textbf{\ding{52}} & 31.28 & 43.74 & 21.05 & 23.24 & 81.21 & 98.01 \\
    \bottomrule[1.5pt]
    \end{tabular}%
  }
  \label{ssy1210:ablationstudy11}%
\end{table}%

\begin{table}[htbp]
\renewcommand\arraystretch{1.0}
  \centering
  \caption{Impact of different components ($pre, en1, en2$) in the transfer experiment cross domains.}
  \scalebox{0.9}{
  \begin{tabular}{ccccccccc}
    \toprule[1.5pt]
    \multicolumn{3}{c}{CS$\rightarrow$ENG} & \multicolumn{2}{c}{Paper-Venue} & \multicolumn{2}{c}{Paper-Field} & \multicolumn{2}{c}{Author ND} \\
    \midrule
    pre & en1 & en2 & {N@10} & {R@10} & {N@10} & {R@10} & {N@10} & {R@10} \\
    \midrule
    \textbf{\ding{56}} & \textbf{\ding{56}} & \textbf{\ding{56}} & 36.48 & 51.95 & 10.95 & 13.32 & 76.96 & 99.49 \\
    \textbf{\ding{52}} & \textbf{\ding{56}} & \textbf{\ding{56}} & 36.60 & 53.67 & 17.43 & 19.99 & 77.25 & 99.49 \\
    \textbf{\ding{52}} & \textbf{\ding{52}} & \textbf{\ding{56}} & 38.84 & 54.88 & 18.06 & 21.10  & 78.96 & 99.49 \\
    \textbf{\ding{52}} & \textbf{\ding{56}} & \textbf{\ding{52}} & 38.73  & 54.34 & 18.16 & 21.04 & 80.19 & 99.49 \\
    \textbf{\ding{52}} & \textbf{\ding{52}} & \textbf{\ding{52}} & 40.27 & 56.08 & 19.50 & 22.51 & 80.32 & 99.61 \\
    \bottomrule[1.5pt]
    \end{tabular}%
  }
  \label{ssy1210:ablationstudy12}%
\end{table}%

\begin{table}[htbp]
\renewcommand\arraystretch{1.0}
  \centering
  \caption{Impact of different types of weights ($\mathbf{p}_{a},\mathbf{p}_{b}$) in the transfer experiment within the same domain.}
  \scalebox{0.93}{
  \begin{tabular}{ccccccccc}
    \toprule[1.5pt]
    \multicolumn{2}{c}{CS$\rightarrow$CS} & \multicolumn{2}{c}{Paper-Venue} & \multicolumn{2}{c}{Paper-Field} & \multicolumn{2}{c}{Author ND} \\
    \midrule
    $\mathbf{p}_{a}$ & $\mathbf{p}_{b}$ & {N@10} & {R@10} & {N@10} & {R@10} & {N@10} & {R@10} \\
    \midrule
    \textbf{\ding{52}} & \textbf{\ding{56}}  & 27.97 & 39.65 & 18.59  & 21.07 & 81.03 & 97.29 \\
    \textbf{\ding{56}} & \textbf{\ding{52}}  & 29.53 & 40.86 & 17.91 & 20.63 & 79.76 & 96.45 \\
    \textbf{\ding{52}} & \textbf{\ding{52}}  & 31.28 & 43.74 & 21.05 & 23.24 & 81.21 & 98.01 \\
    \bottomrule[1.5pt]
    \end{tabular}%
  }
  \label{ssy1210:ablationstudy21}%
\end{table}%

\begin{table}[htbp]
\renewcommand\arraystretch{1.0}
  \centering
  \caption{Impact of different types of weights ($\mathbf{p}_{a},\mathbf{p}_{b}$) in the transfer experiment cross domains.}
  \scalebox{0.9}{
  \begin{tabular}{ccccccccc}
    \toprule[1.5pt]
    \multicolumn{2}{c}{CS$\rightarrow$ENG} & \multicolumn{2}{c}{Paper-Venue} & \multicolumn{2}{c}{Paper-Field} & \multicolumn{2}{c}{Author ND} \\
    \midrule
    $\mathbf{p}_{a}$ & $\mathbf{p}_{b}$ & {N@10} & {R@10} & {N@10} & {R@10} & {N@10} & {R@10} \\
    \midrule
    \textbf{\ding{52}} & \textbf{\ding{56}}  & 39.24 & 54.49 & 16.94 & 19.82 & 80.08 & 99.49 \\
    \textbf{\ding{56}} & \textbf{\ding{52}}  &  39.15 & 54.69 & 18.25 & 21.19 & 78.35 & 99.23 \\
    \textbf{\ding{52}} & \textbf{\ding{52}}  & 40.27 & 56.08 & 19.50 & 22.51 & 80.32 & 99.61 \\
    \bottomrule[1.5pt]
    \end{tabular}%
  }
  \label{ssy1210:ablationstudy22}%
\end{table}%

\subsection{Ablation Study (RQ2)}
In this section, we first investigate the effect of different components proposed in this paper. The abbreviations $\emph{pre}$, $\emph{en1}$, and $\emph{en2}$ in Table~\ref{ssy1210:ablationstudy11} and Table~\ref{ssy1210:ablationstudy12} represent use pre-training strategies, use the query sample side enhancement, and use the positive/negative sample side enhancement, respectively. As shown in Table~\ref{ssy1210:ablationstudy11} and Table~\ref{ssy1210:ablationstudy12}, the performance of the ablated variants is better than that of the base GNN without pre-training. Moreover, the complete PHE performs best in all situations, which means jointly optimizing the two proposed pre-training tasks is necessary. {Then, we investigate the impact of the node-level weight $\mathbf{p}_{a}$ and the type-level weight $\mathbf{p}_{b}$. As shown in Table~\ref{ssy1210:ablationstudy21} and Table~\ref{ssy1210:ablationstudy22}, removing either $\mathbf{p}_{a}$ or $\mathbf{p}_{b}$ can degrade the performance of the algorithm. This is because keeping only $\mathbf{p}_{b}$ will lose individual information while keeping only $\mathbf{p}_{a}$ will lose global information. Removing any one of $\mathbf{p}_{a}$ and $\mathbf{p}_{b}$ will cause the loss of heterogeneous information, so the performance of the algorithm cannot be guaranteed. In other words, keeping only $\mathbf{p}_{a}$ or $\mathbf{p}_{b}$ cannot guarantee that useful information can be obtained, and sometimes even introduces unwanted noise (caused by unreasonable weight estimation).} 

\section{Conclusion}\label{sec:conclusion}
In this paper, we propose the pre-training strategies for the large-scale heterogeneous graph. First, in order to capture high-order structural properties in heterogeneous graphs, we design the contrastive structure-aware pre-training task, where the schema-network subspace is constructed and the attention mechanism is applied to capture fine-grained heterogeneous information. Second, we design the contrastive semantic-aware pre-training task to deal with the problem of semantic mismatch, where a perturbation subspace is constructed to help the model pay attention to semantic neighbors. Extensive experiments on the large-scale heterogeneous graph OAG demonstrate the superiority of the proposed method over its counterparts, especially in the ability to learn transferable knowledge.

\section*{Acknowledgment}
This work was supported by the Early Career Scheme (No. CityU 21219323) and the General Research Fund (No. CityU 11220324) of the University Grants Committee (UGC), the NSFC Young Scientists Fund (No. 9240127), and the Donation for Research Projects (No. 9220187 and No. 9229164).


\bibliographystyle{ACM-Reference-Format}
\bibliography{main}


\end{document}